\documentclass[letterpaper, 10 pt, journal, twoside]{IEEEtran}  
\usepackage[font=small]{caption}
\usepackage[noadjust]{cite}
\def\tablename{Table}
\usepackage{amsmath}
\usepackage{empheq}
\usepackage{multirow}
\usepackage{bm}
\usepackage{graphicx}
\usepackage{amssymb}
\usepackage{svg}
\usepackage{array}
\usepackage{url}
\usepackage{hyperref}
\usepackage{lipsum}
\usepackage[capitalise]{cleveref}
\crefformat{equation}{(#2#1#3)}
\usepackage{tikz}
\usepackage{tikz-3dplot}
\tikzset{small dot/.style={fill=black,circle,outer sep=8pt,scale=0.25}}
\usetikzlibrary{decorations.text}
\usetikzlibrary{shapes.multipart}
\makeatletter

\newif\ifpgfcirclecrosssplitcustomfill
\tikzset{%
	circle cross split part fill/.code=\def\pgf@lib@sh@ccs@list@fill{#1}\pgfcirclecrosssplitcustomfilltrue,%
	circle cross split uses custom fill/.is if=pgfcirclecrosssplitcustomfill}
\pgfdeclareshape{circle cross split}{%
	\nodeparts{text,two,three,four}%
	\savedanchor\centerpoint{%
		\pgfmathsetlength\pgf@xa{\pgfkeysvalueof{/pgf/inner xsep}}%
		\pgfmathsetlength\pgf@ya{\pgfkeysvalueof{/pgf/inner ysep}}%
		\pgf@x\wd\pgfnodeparttextbox
		\pgf@yb\dp\pgfnodeparttextbox
		\pgf@yc\dp\pgfnodeparttwobox
		\ifdim\pgf@yb>\pgf@yc
		\pgf@yc\pgf@yb
		\fi
		\advance\pgf@y-\pgf@yc
		\advance\pgf@x\pgf@xa
		\advance\pgf@y-\pgf@ya
		\advance\pgf@x.5\pgflinewidth
		\advance\pgf@y-.5\pgflinewidth
	}%
	\savedanchor\twoanchor{%
		\pgfmathsetlength\pgf@xa{\pgfkeysvalueof{/pgf/inner xsep}}%
		\pgfmathsetlength\pgf@ya{\pgfkeysvalueof{/pgf/inner ysep}}%
		\advance\pgf@x.5\pgflinewidth
		\advance\pgf@x\pgf@xa
		\advance\pgf@y.5\pgflinewidth
		\advance\pgf@y\pgf@ya
		\pgf@yb\dp\pgfnodeparttextbox
		\pgf@yc\dp\pgfnodeparttwobox
		\ifdim\pgf@yb>\pgf@yc
		\pgf@yc\pgf@yb
		\fi
		\advance\pgf@y\pgf@yc
	}%
	\savedanchor\threeanchor{%
		\pgfmathsetlength\pgf@ya{\pgfkeysvalueof{/pgf/inner ysep}}%
		\pgf@x\wd\pgfnodeparttextbox
		\pgf@yb\dp\pgfnodeparttextbox
		\pgf@yc\dp\pgfnodeparttwobox
		\ifdim\pgf@yb>\pgf@yc
		\pgf@yc\pgf@yb
		\fi
		\advance\pgf@y-\pgf@yc
		\advance\pgf@y-2\pgf@ya
		\advance\pgf@y-\pgflinewidth
		\pgf@yb\ht\pgfnodepartthreebox
		\pgf@yc\ht\pgfnodepartfourbox
		\ifdim\pgf@yb>\pgf@yc
		\pgf@yc\pgf@yb
		\fi
		\advance\pgf@y-\pgf@yc
		\advance\pgf@x-\wd\pgfnodepartthreebox
	}%
	\savedanchor\fouranchor{%
		\pgfmathsetlength\pgf@xa{\pgfkeysvalueof{/pgf/inner xsep}}%
		\advance\pgf@x\wd\pgfnodepartthreebox
		\advance\pgf@x2\pgf@xa
		\advance\pgf@x\pgflinewidth
	}%
	\saveddimen\radius{%
		\pgf@y\ht\pgfnodeparttextbox
		\pgf@yb\ht\pgfnodeparttwobox
		\ifdim\pgf@yb>\pgf@y
		\pgf@y\pgf@yb
		\fi
		\pgf@yc\dp\pgfnodeparttextbox
		\pgf@yb\dp\pgfnodeparttwobox
		\ifdim\pgf@yc>\pgf@yb
		\advance\pgf@y\pgf@yc
		\else
		\advance\pgf@y\pgf@yb
		\fi
		\pgf@yb\ht\pgfnodepartthreebox
		\ifdim\pgf@yb<\ht\pgfnodepartfourbox
		\pgf@yb\ht\pgfnodepartfourbox
		\fi
		\pgf@yc\dp\pgfnodepartthreebox
		\ifdim\pgf@yc<\dp\pgfnodepartfourbox
		\advance\pgf@yb\dp\pgfnodepartfourbox
		\else
		\advance\pgf@yb\pgf@yc
		\fi
		\ifdim\pgf@yc>\pgf@y
		\pgf@y\pgf@yc
		\fi
		\pgfmathsetlength\pgf@ya{\pgfkeysvalueof{/pgf/inner ysep}}%
		\advance\pgf@y2\pgf@ya
		\pgf@x\wd\pgfnodeparttextbox
		\pgf@xa\wd\pgfnodepartthreebox
		\pgf@xb\wd\pgfnodeparttwobox
		\pgf@xc\wd\pgfnodepartfourbox
		\ifdim\pgf@xa>\pgf@x
		\pgf@x\pgf@xa
		\fi
		\ifdim\pgf@xb>\pgf@x
		\pgf@x\pgf@xb
		\fi
		\ifdim\pgf@xc>\pgf@x
		\pgf@x\pgf@xc
		\fi
		\pgfmathsetlength\pgf@xa{\pgfkeysvalueof{/pgf/inner xsep}}%
		\advance\pgf@x2\pgf@xa
		\ifdim\pgf@y>\pgf@x
		\pgf@x\pgf@y
		\fi
		\advance\pgf@x.5\pgflinewidth
		\pgfmathsetlength{\pgf@xb}{\pgfkeysvalueof{/pgf/minimum width}}%
		\pgfmathsetlength{\pgf@yb}{\pgfkeysvalueof{/pgf/minimum height}}%
		\ifdim\pgf@x<.5\pgf@xb
		\pgf@x=.5\pgf@xb
		\fi
		\ifdim\pgf@x<.5\pgf@yb
		\pgf@x=.5\pgf@yb
		\fi
		\pgfmathsetlength{\pgf@xb}{\pgfkeysvalueof{/pgf/outer xsep}}%
		\pgfmathsetlength{\pgf@yb}{\pgfkeysvalueof{/pgf/outer ysep}}%
		\ifdim\pgf@xb<\pgf@yb
		\advance\pgf@x\pgf@yb
		\else
		\advance\pgf@x\pgf@xb
		\fi
	}%
	\inheritanchorborder[from=circle]%
	\inheritanchor[from=circle]{north}%
	\inheritanchor[from=circle]{north west}%
	\inheritanchor[from=circle]{north east}%
	\inheritanchor[from=circle]{center}%
	\inheritanchor[from=circle]{west}%
	\inheritanchor[from=circle]{east}%
	\inheritanchor[from=circle]{mid}%
	\inheritanchor[from=circle]{mid west}%
	\inheritanchor[from=circle]{mid east}%
	\inheritanchor[from=circle]{base}%
	\inheritanchor[from=circle]{base west}%
	\inheritanchor[from=circle]{base east}%
	\inheritanchor[from=circle]{south}%
	\inheritanchor[from=circle]{south west}%
	\inheritanchor[from=circle]{south east}%
	\anchor{two}{\twoanchor}%
	\anchor{three}{\threeanchor}%
	\anchor{four}{\fouranchor}%
	\inheritbackgroundpath[from=circle]
	\beforebackgroundpath{%
		\pgfutil@tempdima=\radius
		\pgfmathsetlength{\pgf@xb}{\pgfkeysvalueof{/pgf/outer xsep}}%
		\pgfmathsetlength{\pgf@yb}{\pgfkeysvalueof{/pgf/outer ysep}}%
		\ifdim\pgf@xb<\pgf@yb
		\advance\pgfutil@tempdima by-\pgf@yb
		\else
		\advance\pgfutil@tempdima by-\pgf@xb
		\fi
		\advance\pgfutil@tempdima by-.5\pgflinewidth%
		\pgfsetshortenstart{0pt}%
		\pgfsetshortenend{0pt}%
		\pgfsetarrows{-}%
		\pgfpathmoveto{\pgfpointadd{\centerpoint}{\pgfqpoint{-\pgfutil@tempdima}{0pt}}}%
		\pgfpathlineto{\pgfpointadd{\centerpoint}{\pgfqpoint{\pgfutil@tempdima}{0pt}}}%
		\pgfpathmoveto{\pgfpointadd{\centerpoint}{\pgfqpoint{0pt}{-\pgfutil@tempdima}}}%
		\pgfpathlineto{\pgfpointadd{\centerpoint}{\pgfqpoint{0pt}{\pgfutil@tempdima}}}%
		\pgfusepathqstroke
	}%
	\behindbackgroundpath{%
		\pgfutil@tempdima=\radius
		\pgfmathsetlength{\pgf@xb}{\pgfkeysvalueof{/pgf/outer xsep}}%
		\pgfmathsetlength{\pgf@yb}{\pgfkeysvalueof{/pgf/outer ysep}}%
		\ifdim\pgf@xb<\pgf@yb
		\advance\pgfutil@tempdima by-\pgf@yb
		\else
		\advance\pgfutil@tempdima by-\pgf@xb
		\fi
		\advance\pgfutil@tempdima by-.5\pgflinewidth%
		\ifpgfcirclecrosssplitcustomfill%
		\pgf@lib@sh@rs@process@list{\pgf@lib@sh@ccs@list@fill}{4}%
		{%
			\pgfmathloop
			\ifnum\pgfmathcounter>4%
			\else%
			\pgf@lib@sh@getalpha\pgf@lib@sh@rs@number{\pgfmathcounter}%
			\edef\pgf@tempa{\csname pgf@lib@sh@rs@\pgf@lib@sh@rs@number @item\endcsname}%
			\ifx\pgf@tempa\pgf@lib@sh@rs@nonetext\else
			\pgfsetfillcolor{\pgf@tempa}%
			\pgf@lib@sh@ccs@angles{\pgfmathcounter}%
			\pgfpathmoveto{\centerpoint}%
			\pgfpathlineto{\pgfpointadd{\centerpoint}{\pgfqpointpolar{\pgf@lib@sh@ccs@angle}{\pgfutil@tempdima}}}%
			\pgfpatharc{\pgf@lib@sh@ccs@angle}{\pgf@lib@sh@ccs@angle@}{\pgfutil@tempdima}%
			\pgfpathclose
			\pgfusepathqfill
			\fi
			\repeatpgfmathloop
		}%
		\fi%
	}%
}
\def\pgf@lib@sh@ccs@angles#1{%
	\ifcase#1\or\def\pgf@lib@sh@ccs@angle{90}%
	\or\def\pgf@lib@sh@ccs@angle{0}%
	\or\def\pgf@lib@sh@ccs@angle{180}%
	\else\def\pgf@lib@sh@ccs@angle{270}%
	\fi
	\edef\pgf@lib@sh@ccs@angle@{\number\numexpr\pgf@lib@sh@ccs@angle+90\relax}%
}
\makeatother

\makeatletter
\def\ps@IEEEtitlepagestyle{%
  \def\@oddfoot{\mycopyrightnotice}%
  \def\@oddhead{\hbox{}\@IEEEheaderstyle\leftmark\hfil\thepage}\relax
  \def\@evenhead{\@IEEEheaderstyle\thepage\hfil\leftmark\hbox{}}\relax
  \def\@evenfoot{}%
}
\def\mycopyrightnotice{%
  \begin{minipage}{\textwidth}
  \centering \scriptsize
\copyright 2021 IEEE.  Personal use of this material is permitted.  Permission from IEEE must be obtained for all other uses, in any current or future media, including reprinting/republishing this material for advertising or promotional purposes, creating new collective works, for resale or redistribution to servers or lists, or reuse of any copyrighted component of this work in other works.
  \end{minipage}
}
\makeatother

\usepackage{subcaption}

\IEEEoverridecommandlockouts                              

\title{\LARGE \bf
A Unified MPC Framework for Whole-Body Dynamic Locomotion and Manipulation
}

\author{Jean-Pierre Sleiman, Farbod Farshidian, Maria Vittoria Minniti, Marco Hutter
\thanks{Manuscript received: October, 15, 2020; Revised January, 16, 2021; Accepted February 11, 2021}
\thanks{This paper was recommended for publication by
Editor Abderrahmane Kheddar upon evaluation of the Associate Editor and
Reviewers’ comments.}
\thanks{This research was supported in part by the Swiss National Science Foundation through the National Centre of Competence in Research Robotics (NCCR Robotics) and in Research Digital Fabrication (NCCR Dfab), and in part by TenneT.}
\thanks{All authors are with the Robotic Systems Lab, ETH Zurich, Zurich 8092, Switzerland. 
(Email: {\tt\small  jsleiman@ethz.ch})
}
\thanks{Digital Object Identifier (DOI): see top of this page.}
}

\begin{document}

\maketitle

\begin{abstract}
In this paper, we propose a whole-body planning framework that unifies dynamic locomotion and manipulation tasks by formulating a single multi-contact optimal control problem. We model the hybrid nature of a generic multi-limbed mobile manipulator as a switched system, and introduce a set of constraints that can encode any pre-defined gait sequence or manipulation schedule in the formulation. Since the system is designed to actively manipulate its environment, the equations of motion are composed by augmenting the robot's centroidal dynamics with the manipulated-object dynamics. This allows us to describe any high-level task in the same cost/constraint function. The resulting planning framework could be solved on the robot's onboard computer in real-time within a model predictive control scheme. This is demonstrated in a set of real hardware experiments done in free-motion, such as base or end-effector pose tracking, and while pushing/pulling a heavy resistive door. Robustness against model mismatches and external disturbances is also verified during these test cases.    
\end{abstract}
\begin{IEEEkeywords}
Multi-Contact Whole-Body Motion Planning and Control, Mobile Manipulation, Legged Robots, Optimization and Optimal Control.
\end{IEEEkeywords}

\section{INTRODUCTION} \label{Introduction}
\IEEEPARstart{W}{e} often judge the agility of a poly-articulated robotic system, such as a humanoid or quadruped, by the degree to which it is able to mimic its biological counterpart. This resemblance should appear in the robot's ability to properly coordinate a wide range of complex body movements, and its ability to effectively interact with its environment. Such an interaction could be directed towards moving and balancing the robot's base (locomotion), or towards moving another object (manipulation). The governing dynamics in such problems are hybrid, underactuated, and highly non-linear; this in turn renders the design of controllers for such systems quite challenging.

A broad range of work in the literature relies on a decomposition of the full control problem into two main units, namely a planning module and a tracking module. The latter is responsible for generating the torque commands needed to compliantly track the high-level references computed by the planner. Typically, the tracking controller is based on variants of the standard operational-space inverse dynamics approach \cite{Khatib}. These variants formulate the tracking problem within an optimization setting to properly resolve the system's redundancy while allowing for the incorporation of system constraints \cite{Kanoun,HAL_Hierarchical,Scott,WBC2,Fahmi}. As for the planning module, which is also the main focus of this paper, it is responsible for generating center-of-mass (CoM) motions, contact locations and forces, as well as limb motion trajectories. These can be computed either simultaneously within the same planner, or separately in decoupled sub-modules, depending on the adopted dynamic model. Generally, these planners rely on simplified template models -- with a few notable exceptions that use the full dynamics \cite{Neunert,HRP2,Posa,Crocoddyl} -- where a wide spectrum of options trades off physical accuracy against computational complexity. For instance, Bellicoso et al. \cite{WBC1} demonstrate agile dynamic locomotion on a quadrupedal robot by decomposing their planning framework into the aforementioned three main elements. The CoM trajectories are generated and updated online on the basis of the zero-moment-point (ZMP) dynamic stability criterion \cite{Sardain}. On the other hand, in the works of Di Carlo et al. \cite{MitCheetah} and Villareal et al. \cite{HyQ}, the quadruped is modeled as a single rigid-body subject to contact patches. With certain assumptions made on the base's orientation and angular velocity, they are able to formulate their planning problem as a convex optimization problem. This is then solved within a Model Predictive Control (MPC) scheme to compute online CoM motions and reference contact forces.
\begin{figure}[t]
    \centering
                \scalebox{0.75}{
       \begin{tikzpicture}
\node (A) at (0,0) {\includegraphics[trim={0cm 7cm 0cm 9cm},clip,scale=0.1,width = 3.5cm, keepaspectratio]{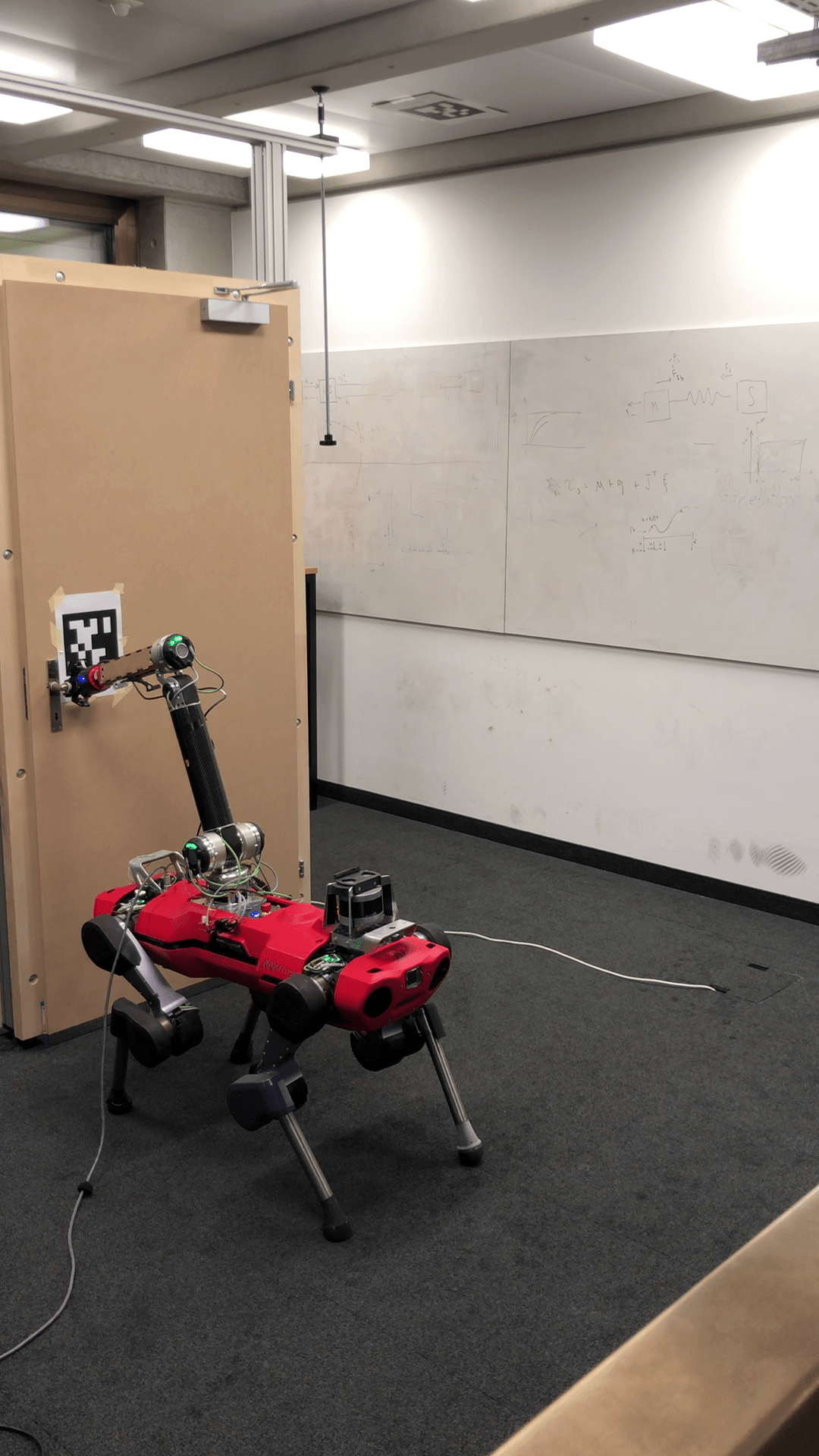}};
\node[right] (B) at (A.east) {\includegraphics[trim={0cm 7cm 0cm 9cm},clip,scale=0.1,width = 3.5cm, keepaspectratio]{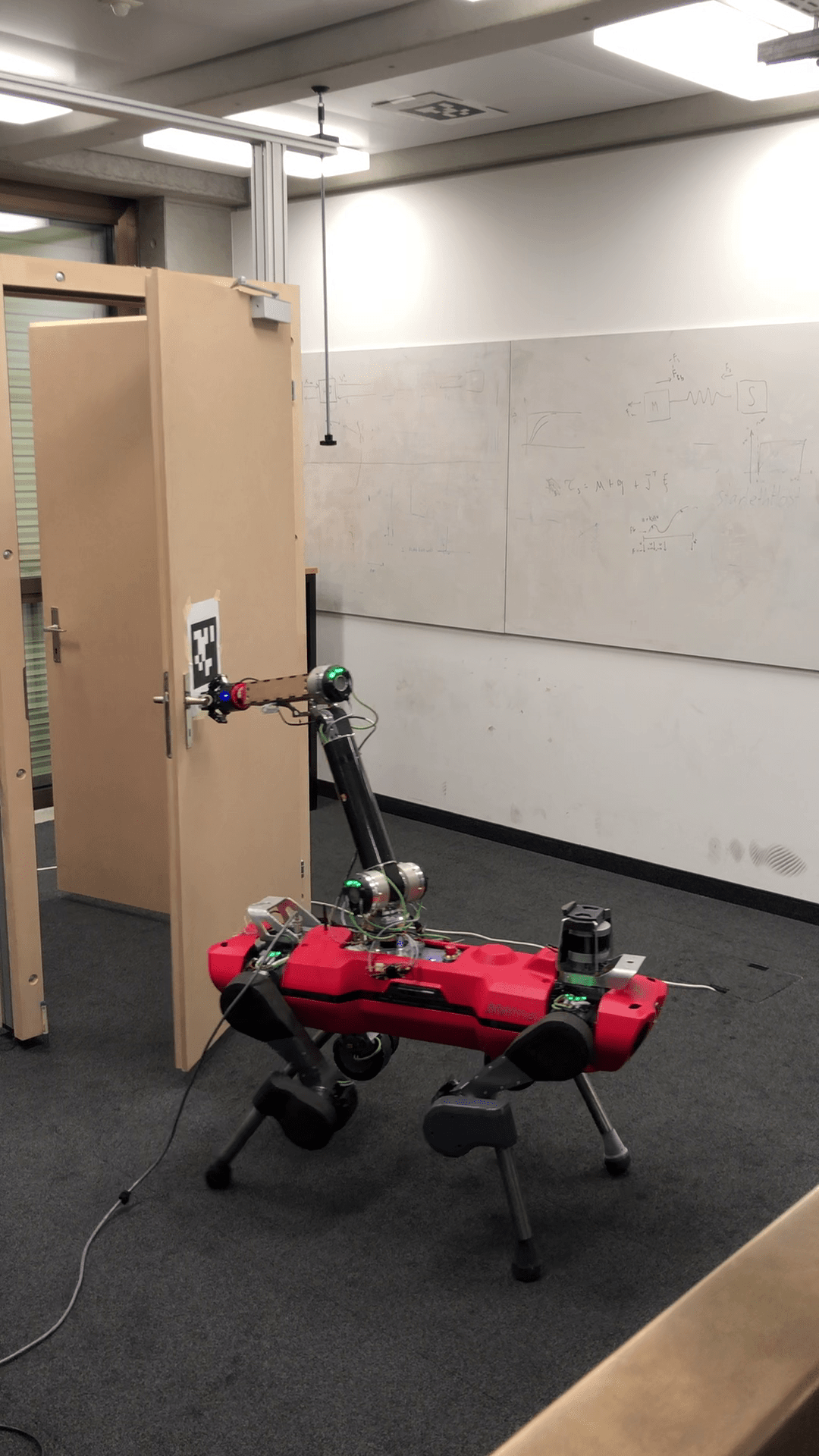}};
\node[right] (C) at (B.east) {\includegraphics[trim={0cm 7cm 0cm 9cm},clip,scale=0.1,width = 3.5cm, keepaspectratio]{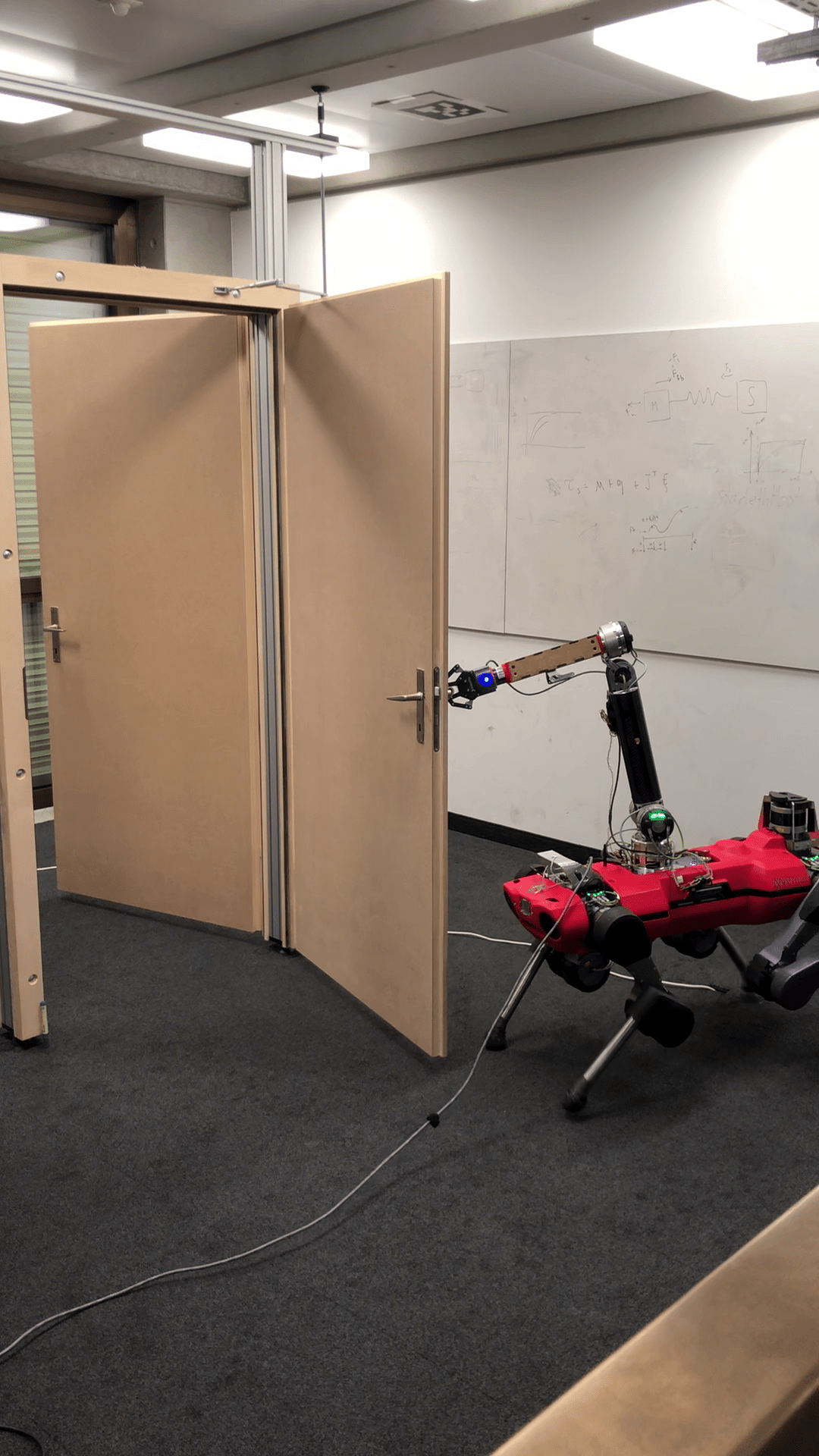}};
\node(AA) at (A.north west) {};
\node [fill=white, draw = black, minimum height = 0.3 cm, minimum width = 0.4cm, xshift = 0.675cm,yshift = -0.25cm, anchor = north east] at (AA) {\small 1};
\node(AA) at (B.north west) {};
\node [fill=white,  draw = black, minimum height = 0.3 cm, minimum width = 0.4cm, xshift = 0.675cm,yshift = -0.25cm, anchor = north east] at (AA) {\small 2};
\node(AA) at (C.north west) {};
\node [ fill=white, draw = black, minimum height = 0.3 cm, minimum width = 0.4cm, xshift = 0.675cm,yshift = -0.25cm, anchor = north east] at (AA) {\small 3};
\end{tikzpicture}}
    \caption{A quadrupedal mobile manipulator,  consisting of an ANYmal C platform equipped with a custom-made robotic arm, pulling a heavy resistive door open.}
    \label{fig:AlmaCDoorPulling}
\end{figure}
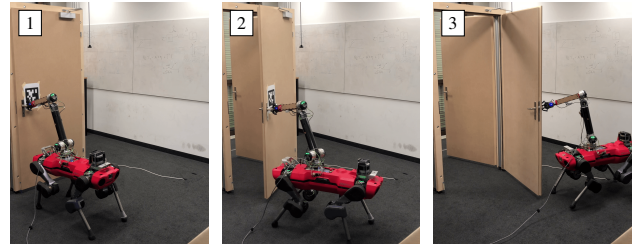
The same template model is also adopted in \cite{Winkler}, where a single trajectory optimization (TO) framework is used to compute contact-schedules, swing-leg motions and base trajectories, for locomotion over uneven terrain. Dai et al. \cite{Dai} use a full centroidal dynamic description \cite{Orin} along with the full kinematics of a humanoid robot within a trajectory optimization setting. Their formulation is also able to discover contacts automatically through the use of complementarity constraints \cite{Posa}. A similar model is also adopted in \cite{kudruss}, where the authors demonstrate real-hardware experiments on a humanoid robot climbing stairs while grabbing a hand-rail, given a predefined contact sequence. However, all of these formulations could not be solved in real-time, and thus do not allow for fast online-replanning. On the other hand, there has been relatively more recent attempts to have a unified locomotion MPC-scheme that takes advantage of fast trajectory-optimization techniques. One such example is presented in \cite{Ruben1}, where the multi-contact problem is formulated within the framework introduced in \cite{Farbod1,Farbod2}. This work makes use of a similar optimal control problem description. 

When it comes to articulated whole-body manipulation, the most notable contributions could be found in the works of Murphy et al. \cite{BostonDynamics}, where a quadrupedal mobile manipulator is shown performing coordinated athletic motions in a dynamic lifting and throwing task, and in Bellicoso et al. \cite{Alma} where a similar system is used in a door pushing scenario. However, unlike this work, in \cite{BostonDynamics} the manipulation task is planned offline and is separated from the locomotion planner; while in \cite{Alma}, the door-opening task is solved in a reactive fashion by explicitly specifying gripper forces to be tracked by the whole-body controller.   
    

The main contributions of this work are three-fold: 
\begin{itemize}
    \item We exploit the duality between dynamic locomotion and manipulation tasks to formulate a unifying multi-contact optimal control problem. To this end, we adopt a switched-systems perspective when handling contact-making and contact-breaking events. Moreover, we use a single set of constraints to describe any pre-defined gait sequence or manipulation contact-schedule. 
    \item We propose an augmented model consisting of the manipulated-object dynamics, the robot's centroidal dynamics, and the full kinematics. This description enables us to encode any robot-centric or object-centric task in the same cost/constraint function.
    \item We show that the resulting OC framework can resolve a wide variety of free-motion tasks and manipulation problems. By exploiting the whole-body natural dynamics, the planner is able to generate coordinated maneuvers that push our system's performance limits without violating any physical constraints. Most importantly, we manage to solve the OC problem in real-time, in a receding-horizon fashion. This results in an MPC scheme that can easily be deployed on hardware and solved with limited on-board computational power. 
\end{itemize}
To the best of our knowledge, this work is one of the first examples demonstrating a real-hardware application of a whole-body MPC framework that unifies dynamic locomotion and manipulation.  

\section{PROBLEM FORMULATION} \label{Formulation}
\subsection{Whole-Body Planner} \label{WholeBodyPlanner}
The nonlinear-MPC framework adopted in this work is based on the solver introduced in \cite{Farbod1,Farbod2} which is tailored to handle optimal control problems involving hybrid dynamical systems, by treating them as switched systems with predefined modes. It also offers the possibility to optimize for the switching-times, together with the optimal state and input trajectories. The underlying core algorithm is the Sequential-Linear-Quadratic (SLQ) technique, a continuous-time version of the iterative-LQR (iLQR) method \cite{ilqr2}. More specifically, SLQ can be classified as a variant of Differential Dynamic Programming \cite{Mayne}, an indirect trajectory optimization approach that relies on the following mechanism: Given a nominal state and input trajectory, perform a forward rollout of the dynamics, compute the quadratic approximations of the cost function and dynamics (SLQ uses a first-order approximation of the dynamics instead), then perform a backward pass on the resulting Riccati equations to finally compute a control policy consisting of a feedforward and feedback term. The full sequence of computations (i.e., one SLQ iteration) has a complexity that is linear with respect to the time horizon, unlike direct TO methods which have cubic complexity.

Furthermore, the unconstrained-SLQ algorithm has been extended to handle state-input and state-only equality constraints \cite{Farbod1}, as well as inequality path constraints \cite{Ruben1}, through projections, penalty functions and barrier functions, respectively. Hence, we are able to devise the following optimal control problem
\begin{equation} \label{eq:OCProblem}
    \begin{cases}
    \underset{\bm u(.)}{\min} \ \ \Phi(\bm x(T)) + \displaystyle \int_{0}^{T} L(\bm x(t), \bm u(t), t)dt \\[2ex]
    \text{s.t.} \ \ \dot{\bm{x}}(t) = \bm f(\bm x(t), \bm u(t), t) \\
    \ \ \ \ \ \bm g_1(\bm x(t), \bm u(t), t) = 0 \\
    \ \ \ \ \ \bm g_2(\bm x(t), t) = 0 \\
    \ \ \ \ \ \bm h(\bm x(t), \bm u(t), t) \geq 0 \\
    \ \ \ \ \ \bm x(0) = \bm x_0,
    \end{cases}
\end{equation} 
where $\bm x(t) \in \mathbb{R}^{n_x}$ and $\bm u(t) \in \mathbb{R}^{n_u}$ are vectors of state and input variables, while $L(\bm x, \bm u, t)$ and $\Phi(\bm x(T))$ are the stage cost (Lagrange term) and terminal cost (Mayer term), respectively. This problem is solved in closed-loop with the SLQ-MPC framework that essentially runs the constrained-SLQ algorithm in a receding horizon fashion within a real-time iteration scheme \cite{Diehl}.
To simplify notation, we omit any dependence of~\cref{eq:OCProblem} on the modes and their corresponding switching-times, which are assumed to be fixed in our case. It is worth mentioning that the switched nature of our dynamic locomotion/manipulation setup is captured in the constraints rather than in the system dynamics. Therefore, this matter will be made clear in ~\cref{EqualityConstraints,InequalityConstraints}, while the discussion regarding cost function design is deferred to~\cref{Experiments}.

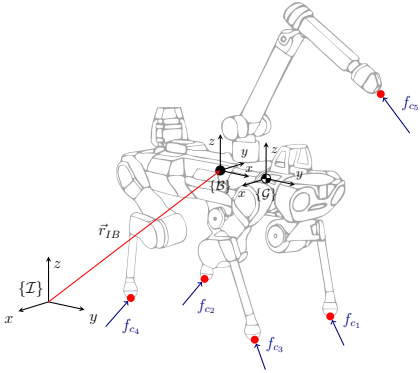
\begin{figure}[t]
    \centering
    \vspace*{2mm}
\tdplotsetmaincoords{75}{130}
\scalebox{0.5}{
\begin{tikzpicture}[tdplot_main_coords]

\node[opacity =0.35] at (-10,0.25,0) {\includegraphics[scale=0.35]{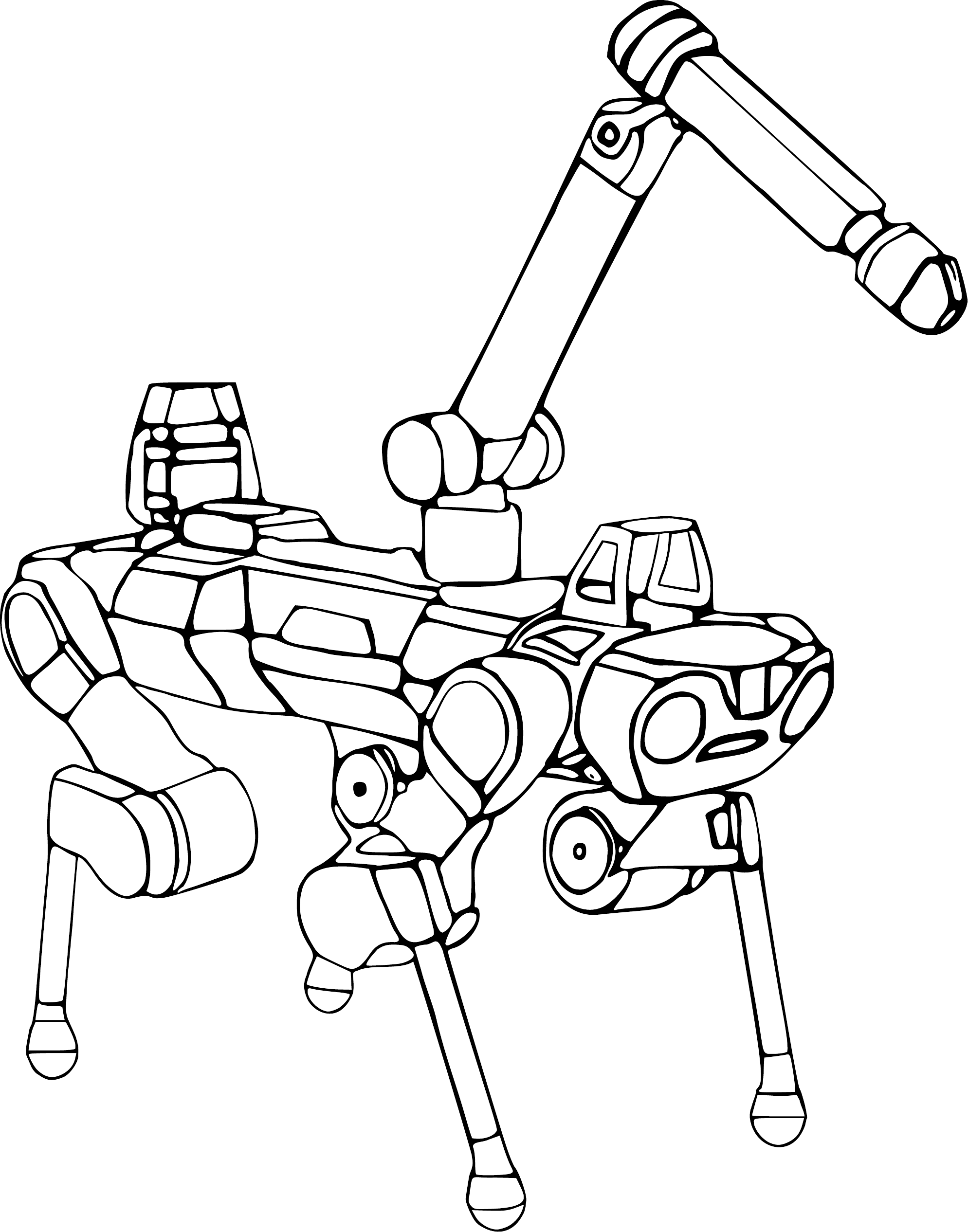}};

\tdplotsetrotatedcoords{0}{0}{0}
\coordinate (shift) at (-2,0,-2) ;
\tdplotsetrotatedcoordsorigin{(shift)}
\draw[tdplot_rotated_coords,thick,-stealth] (0,0,0)  -- (1.25,0,0) node[anchor=north east]{\large $x$};
\draw[tdplot_rotated_coords,thick,-stealth] (0,0,0)  --  (0,1.25,0) node[anchor=north west]{\large $y$};
\draw[tdplot_rotated_coords,thick,-stealth] (0,0,0)  --  (0,0,1.25) node[anchor=north west]{\large$z$};
\node[anchor = south east ] at (-2,0,-2)  {\large $\{ \mathcal{I} \}$};

\tdplotsetrotatedcoords{0}{0}{0}
\coordinate (shift) at (-10,-0.75,-0.15) ;
\tdplotsetrotatedcoordsorigin{(shift)}
\draw[tdplot_rotated_coords,thick,-stealth,black] (0,0,0)  --(-1,0,0) node[above]{$y$};
\draw[tdplot_rotated_coords,thick,-stealth,black] (0,0,0)  -- (0,1,0) node[above]{$x$};
\draw[tdplot_rotated_coords,thick,-stealth,black] (0,0,0) -- (0,0,1) node[anchor=north east]{ $z$};
\node[anchor = north,black, yshift= -0.125 cm] at (-10,-0.75,-0.15) {$\{ \mathcal{B} \}$};

\tdplotsetrotatedcoords{0}{0}{0}
\coordinate (shift) at (-11,0,-0.45) ;
\tdplotsetrotatedcoordsorigin{(shift)}
\draw[tdplot_rotated_coords,thick,-stealth,black] (0,0,0)  --(1,0,0) node[below]{$x$};
\draw[tdplot_rotated_coords,thick,-stealth,black] (0,0,0)  -- (0,1,0) node[above,xshift=0.125cm]{ $y$};
\draw[tdplot_rotated_coords,thick,-stealth,black] (0,0,0) -- (0,0,1) node[anchor=north west]{ $z$};
\node[anchor = north,black, yshift= -0.125 cm] at (-11,0,-0.45) {$\{ \mathcal{G} \}$};

\node[] (C) at (-10,-0.75,-0.15) {
	\begin{tikzpicture}
	\node [scale=0.5,
	circle cross split,
	draw,
	circle cross split part fill={black,black,black,black},
	] (s) {};
	\end{tikzpicture}
};

\node[] (C) at (-11,0.0,-0.45) {
	\begin{tikzpicture}
	\node [scale=0.5,
	circle cross split,
	draw,
	circle cross split part fill={black,white,white,black},
	] (s) {};
	\end{tikzpicture}
};
\draw[thick,-stealth,red](-2,0,-2) -- (-10,-0.75,-0.15) node[pos=0.45, anchor = south east,black] {\large $\vec{r}_{IB}$};

\node[circle,draw=none,fill=red, inner sep=0pt,minimum size=6pt] (a) at (-15.325,0.35,1.05) {};
\draw[thick,->,blue!50!black]  (-15.75,1,0) -- (a) node[pos= 0.5,anchor = south west] {\large $f_{c_{5}}$};

\node[circle,draw=none,fill=red, inner sep=0pt,minimum size=6pt] (a) at (-14.25,-.5,-5.)  {};
\draw[thick,->,blue!50!black]  (-13.625,0.5,-5.5) -- (a)  node[pos= 0.5,anchor = south west] {\large $f_{c_{1}}$};

\node[circle,draw=none,fill=red, inner sep=0pt,minimum size=6pt] (a) at  (-11.125,-.5,-5)   {};
\draw[thick,->,blue!50!black]  (-10.375,0.5,-5.5) -- (a)  node[pos= 0.5,anchor = south west] {\large $f_{c_{3}}$};

\node[circle,draw=none,fill=red, inner sep=0pt,minimum size=6pt] (a) at(-6,-.5,-2.8)  {};
\draw[thick,->,blue!50!black]  (-6.75,-2,-4) -- (a)  node[pos= 0.5,anchor = north west] {\large $f_{c_{4}}$};

\node[circle,draw=none,fill=red, inner sep=0pt,minimum size=6pt] (a) at (-9.05,-.5,-2.9) {};
\draw[thick,->,blue!50!black]  (-7.25,0.25,-3.125) -- (a)  node[pos= 0.5,anchor = north west] {\large $f_{c_{2}}$};

\end{tikzpicture}}
    \caption{Illustration of the multi-limbed floating-base system used in this work. The sketch depicts the main reference frames used in our derivation of the equations of motion (inertial $\{\mathcal{I}\}$, base $\{\mathcal{B}\}$, and centroidal $\{\mathcal{G}\}$ frames), in addition to the contact forces acting on the limbs.}
    \label{fig:AlmaWireFrame}
    \vspace{0mm}
\end{figure}
\subsubsection{System Modeling}
A poly-articulated floating-base system, such as the one depicted in~\cref{fig:AlmaWireFrame}, can be properly modeled as an unactuated 3D rigid body to which is attached a set of fully-actuated limbs. The resulting dynamics are therefore governed by the following set of equations 
\begin{subequations}
\begin{align} \label{eq:Unactuated}
    \bm M_u(\bm q) \dot{\bm \nu} + \bm b_u(\bm q,\bm \nu) &= \bm J_{c_u}^T(\bm q) \bm F_c \\
    \bm M_a(\bm q) \dot{\bm \nu} + \bm b_a(\bm q, \bm \nu) &= \bm \tau_a + \bm J_{c_a}^T(\bm q) \bm F_c
\end{align}
\end{subequations} 
where $\bm q, \bm \nu \in \mathbb{R}^{6 + n_a}$ are the generalized coordinates and generalized velocities, respectively. A ZYX-Euler angle parameterization is assumed to represent the base's orientation. $\bm M$ is the generalized mass matrix, $\bm b$ represents the nonlinear effects (i.e., Coriolis, centrifugal, and gravitational terms), and $\bm \tau_a$ is the vector of actuation torques. $\bm J_c$ is a matrix of stacked contact Jacobians, while $\bm F_c$ is a vector of stacked contact wrenches. The subscripts $u$ and $a$ correspond to the unactuated and actuated parts of the defined quantities, respectively. 
Under the mild assumption that one has sufficient control authority in the robot's joints, it would be justifiable to independently consider subsystem~\cref{eq:Unactuated} in the TO formulation as a simplified template model. In fact, with the proper transformation applied to~\cref{eq:Unactuated}, one could equivalently retrieve the Newton-Euler equations applied at the robot's center of mass (CoM), or the centroidal dynamics \cite{Wieber2006}
\begin{equation} \label{eq:CentroidalDynamics}
    \dot{\bm h}_{com} = 
    \begin{bmatrix}\sum\limits_{i = 1}^{n_c} \bm f_{c_i} + m \bm g \\
    \sum\limits_{i = 1}^{n_c} \bm r_{com, c_i} \times \bm f_{c_i} + \bm \tau_{c_i}
    \end{bmatrix}.
\end{equation} \\
$\bm h_{com} = (\bm p_{com}, \bm l_{com}) \in \mathbb{R}^{6}$ is defined as the centroidal momentum, with $\bm p_{com}$ and $\bm l_{com}$ being the linear momentum and angular momentum about the centroidal frame $\mathcal{G}$\footnote{The centroidal frame is a reference frame attached to the robot's center of mass and aligned with the world frame.}, respectively. $\bm r_{com, c_{i}}$ denotes the position of the contact point $c_i$ with respect to the center of mass, while $\bm f_{c_i}$ and $\bm \tau_{c_i}$ are the contact forces and torques applied by the environment on the robot at $c_i$.

In order to capture the effect of the generalized coordinates' rate of change on the centroidal momentum, we consider the mapping introduced in \cite{Orin} through the centroidal momentum matrix (CMM) $\bm A(\bm q) \in \mathbb{R}^{6\times(6+n_a)}$ - which is constructed as a function of the system's full kinematic configuration and the multi-body inertias-
\begin{equation} \label{eq:centroidaldynamics2}
    \bm h_{com} = \underbrace{[\bm A_{b}(\bm q) \ \ \bm A_j(\bm q)]}_{\bm A(\bm q)} \begin{bmatrix} \dot{\bm q}_b \\ \dot{\bm q}_j\end{bmatrix}.
\end{equation} 
This could be rearranged in the following form
\begin{equation} \label{eq:BaseDerivative}
    \dot{\bm q}_b = \bm A^{-1}_b \left(\bm h_{com} - \bm A_j \dot{\bm q}_j \right),  
\end{equation}
where $\bm q_b = (\bm r_{IB}, \ \bm \Phi^{zyx}_{IB}) \in \mathbb{R}^6$ is the base pose with respect to the inertial frame. It is important to note that including~\cref{eq:BaseDerivative} in our equations of motion rids our model of the standard massless-limbs assumption, which is inherent to most template models of legged systems. Having laid down the foundations for defining physically consistent dynamics in our motion planner, we are now able to represent the robot as a dynamical system with state vector ${\bm x_r = (\bm h_{com}, \ \bm q_b, \ \bm q_j) \in \mathbb{R}^{12 + n_a}}$ and input vector ${\bm u = (\bm f_{c_1}, \ ..., \ \bm f_{c_{n_c}}, \ \bm v_j) \in \mathbb{R}^{3 n_c + n_a}}$ with 
\begin{equation} \label{eq:JointDerivative}
    \dot{\bm q}_j = \bm v_j,
\end{equation}
here we have neglected contact torques as inputs by assuming points of contact rather than patches. The system of interest, illustrated in~\cref{fig:AlmaWireFrame}, consists of 16 actuated joints (3 per leg and 4 for the arm) and 5 potential contact points; this entails a total of 28 state variables and 31 inputs. 

When dealing with whole-body manipulation problems, having a planner that directly encapsulates the task description and is also aware of the robot-object dynamic coupling is fundamental for effective task-attainment. This could be achieved by augmenting the object state ${\bm x_o = (\bm q_o, \bm v_o) \in \mathbb{R}^{2n_o}}$ to $\bm x_r$, where the object dynamical system is defined similarly to~\cref{eq:Unactuated}
\begin{equation} \label{eq:ObjectDynamics}
    \dot{\bm x}_o = \begin{bmatrix} \bm v_o \\[1ex] \bm M^{-1}_o \left(-\bm J^T_{c_o} \bm f_{c_5} - \bm b_o \right)\end{bmatrix}.
\end{equation}
The term $\bm b_0$ captures all position and velocity-dependent generalized forces, such as spring-damper effects. We note that the underlying assumptions that make such a state augmentation possible are that the object model structure and parameters are known, and that the object state is observable as it needs to be continuously fed back to the MPC solver. 
Finally, the full system flow-map $\dot{\bm x} = \bm f(\bm x, \bm u)$ with the augmented state ${\bm x = (\bm x_r, \bm x_o)}$ could be set up by collecting~\cref{eq:CentroidalDynamics,eq:BaseDerivative,eq:JointDerivative,,eq:ObjectDynamics}. \newline

\subsubsection{Equality Constraints} \label{EqualityConstraints}
All of the equality constraints are defined at the level of the potential contact points, which could be in one of two states: open or closed. Thereby, the constraints would depend on a predefined mode-schedule that consists of a mode-sequence coupled with a set of switching times. For the sake of compactness in notation, we denote the set of all closed contacts by $\mathcal{C}$, the set of feet contact points by $\mathcal{F}$, and the set of arm contacts by $\mathcal{A}$ (which in our case is a singleton). Given a fixed mode instance $\mathcal{C}_j$ that starts at time $s^0_j$ and ends at $s^f_j$, the state-input equality constraints corresponding to this mode ${\bm g_{1_j}(\bm x, \bm u, t)}$ can be established as the following:
$\forall t \in [s_j^0, \ s_j^f]$ and $\forall i \in \{1, ..., n_c\}$
%
\begin{subequations}\label{eq:equalityConstraints}
    \begin{empheq}[left={\empheqlbrace\,}]{align}
&\bm v_{c_i} = \bm 0 &&\text{if } c_i \in \mathcal{C}_j \ \wedge \ c_i \in \mathcal{F} \label{eq:eight1}
\\
&\bm v_{c_i} \cdot \hat{\bm n} = v^*(t) &&\text{if } c_i \in \bar{\mathcal{C}}_j \ \wedge \ c_i \in \mathcal{F} \label{eq:eight2}
\\
&\bm v_{c_i} - \bm J_{c_o} \bm v_o = 0 &&\text{if }  c_i \in \mathcal{C}_j \ \wedge \ c_i \in \mathcal{A} \label{eq:eight3}
\\
&\bm f_{c_i} = \bm 0 &&\text{if } c_i \in \bar{\mathcal{C}}_j,\label{eq:eight4}
   \end{empheq}
\end{subequations}
where $\bm v_{c_i}$ is the absolute linear velocity of contact point $c_i$ expressed in the inertial frame. The constraints presented in~\cref{eq:equalityConstraints} have the following implications: forces at open contacts vanish~\cref{eq:eight4}, the foot of a stance leg should not separate or slip with respect to the ground~\cref{eq:eight1}, and a grasped object should remain in contact with the end-effector at the gripping point~\cref{eq:eight3}. Furthermore, all swing legs should track a reference trajectory ${v^*(t)}$ along the surface normal $\hat{\bm n}$~\cref{eq:eight2}, thus leaving the remaining orthogonal directions -- which determine the stride-length -- free for the planner to resolve. It is worth noting that while it is true that a wide variety of manipulation tasks could be handled by supposing a continuous grasp, allowing for an open arm-contact generalizes the planning framework by additionally covering non-prehensile tasks. In such cases, the robot is expected to lose contact with the object during the manipulation period. 
\newline 
\subsubsection{Inequality Constraints} \label{InequalityConstraints}
One of the downsides of using a DDP-based method for trajectory optimization is that inequality constraints are not naturally handled by Riccati-based solvers. As mentioned in the introduction of~\cref{WholeBodyPlanner}, one remedy has been proposed in \cite{Ruben1} which extends the original SLQ formulation by augmenting the inequalities to the cost in the form of a relaxed log-barrier function. In a separate concurrent paper \cite{InequalitySLQ}, we highlight the issues that could arise from such a method, and we propose a novel inequality-constrained SLQ-MPC algorithm based on an augmented-Lagrangian formulation \cite{Nocedal}. This algorithm is applied in this work to properly handle inequality constraints that appear in the planner, while state-input equalities are still handled with the projection technique introduced in \cite{Farbod1}. Briefly, the idea would be to transform the inequality constrained problem into an unconstrained one by constructing the augmented Lagrangian function $\mathcal{L}_A$. Then at each iteration, $\mathcal{L}_A$ is minimized with respect to the primal variables $\bm x(t)$ and $\bm u(t)$ through a single call to the equality-constrained SLQ loop
\begin{equation}
    (\bm x^*_{k+1}, \ \bm u^*_{k+1}) = \underset{\bm x_k, \bm u_k}{\text{argmin }} \mathcal{L}_A(\bm h(\bm x_k, \bm u_k), \bm \lambda_k),
\end{equation}
while the Lagrange dual function is maximized in the direction of the dual variables $\bm \lambda(t)$ through an update rule
\begin{equation}
    \bm \lambda^*_{k+1} = \bm \Pi\big(\bm \lambda_k, \bm h(\bm x^*_{k+1}, \bm u^*_{k+1}) \big).
\end{equation}
The algorithm eventually converges to a KKT point $(\bm x^*, \bm u^*, \bm \lambda^*)$ which is a potential primal-dual optimum. 

To ensure that our planner generates dynamically feasible motions and forces that would also respect the system's intrinsic operational limits, we introduce the following set of constraints:
$\forall t \in [s_j^0, \ s_j^f]$ and $\forall i \in \{1, ..., n_c\}$
\begin{equation}\label{eq:inequalityConstraints}
\left\{
\begin{split}
&-\bm v_{j_{max}} \leq \bm v^{arm}_j \leq \bm v_{j_{max}} & 
\\[0.5ex]
&-\bm \tau_{max} \leq \bm J^{a^T}_{c_i} \bm f_{c_i} + \bm g^a \leq \bm \tau_{max} & \text{if }  c_i \in \mathcal{C}_j \ \wedge \ c_i \in \mathcal{A}
\\[0.5ex]
& \mu_s f^z_{c_i} - \sqrt{f^{x^2}_{c_i} + f^{y^2}_{c_i} + \epsilon^2} \geq 0 & \text{if } c_i \in \mathcal{C}_j \ \wedge \ c_i \in \mathcal{F}.
\end{split}
\right.
\end{equation}
where $\bm J^{a}$ and $\bm g^a$ are the arm Jacobian and generalized gravitational torques, respectively. These inequalities guarantee that the arm's joint velocity and torque limits are not violated, and that the feet contact forces remain inside the friction cone with friction coefficient $\mu_s$ ($\epsilon \neq 0$ is needed to smoothen the constraint). We note that the torque constraint encompasses the dynamic effects of the object in $\bm f_{c_i}$, but neglects the arm dynamics: Inertial terms are omitted since joint accelerations are not accessible in the planner, while velocity-dependent terms are neglected to avoid reductions in the MPC frequency. Since the commanded torques are not computed directly in the MPC layer but rather in the tracking controller, we found this to be a safe and reasonable assumption. Strict feasibility with respect to the real torque limits is indeed imposed in the QP-based controller of~\cref{sec:WBC}.  
\newline
\subsubsection{Complementary Remarks}
In order to compute the necessary kinematic transforms, Jacobians and CMM of our robot, we rely on the fast implementation of rigid-body dynamic algorithms provided by the \textit{Pinocchio} C++ library \cite{pinocchio1, pinocchio2}. 
We also recall that the SLQ algorithm requires the gradients of the dynamics and constraints to perform the linear-quadratic approximation of~\cref{eq:OCProblem}. Accordingly, we carry out a basic comparison where we evaluate an implementation based on automatic differentiation with CppAD \cite{cppad}, against an implementation based on derived analytical expressions\footnote{These derivations essentially make use of \textit{Pinocchio}'s computation of the centroidal dynamics and their analytical derivatives.}, of which the former turns out to be slightly more efficient.  
\subsection{Whole-Body Controller} \label{sec:WBC}
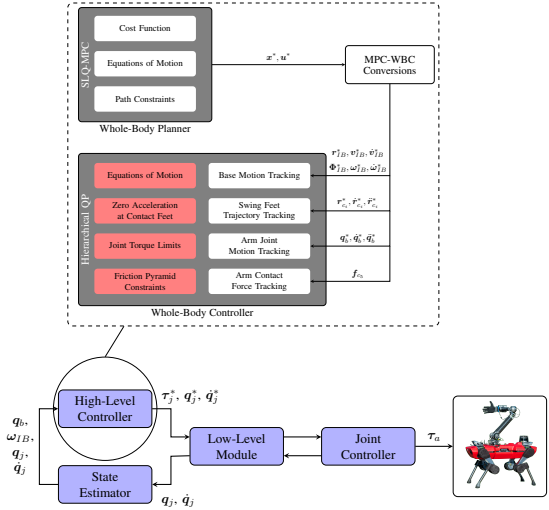
\begin{figure}
\vspace*{4mm}
    \centering
    \scalebox{0.5}{
\begin{tikzpicture}
 \node [draw = black,fill = none, dashed, rounded corners=0.125 cm]  (A) at (0,0) {\scalebox{0.675}{
\begin{tikzpicture}
\node [solid, draw = black,fill = black!50, minimum height = 4.5 cm, minimum width = 5.25 cm,text width = 3.5 cm,rounded corners=0.125 cm] (a) at (-1.875,-0.25) {};
\node[below,rotate = 90,white] at (a.west) {\large SLQ-MPC};
\node [solid, fill = white, minimum height = 1 cm, minimum width = 4 cm,rounded corners=0.125 cm, yshift= -0.325cm,below] (b) at (a.north) {Cost Function};
\node [solid, align = center, fill = white, minimum height = 1 cm, text width = 3.5 cm, minimum width = 4 cm,rounded corners=0.125 cm, yshift= -0.375 cm,below] (c) at (b.south) {Equations of Motion};
\node [solid, align = center, fill = white, minimum height = 1 cm, text width = 3.5 cm, minimum width = 4 cm,rounded corners=0.125 cm,yshift= -0.375 cm,below] (d) at (c.south) {Path Constraints};
\node [solid, align = center, draw=black,fill = none, minimum height = 1.5 cm, minimum width =3.5 cm,text width = 3 cm,rounded corners=0.125 cm,right,xshift = 5.25 cm] (e) at (a.east) {\large MPC-WBC\\ Conversions};
\draw[solid,thick,-stealth] (a) -- (e) node [pos = 0.5,above] {$\bm x^*, \bm u^*$};
\node [solid, draw = none,fill = none, minimum height = 0.5 cm, minimum width = 5.5 cm,text width = 4 cm,rounded corners=0.125 cm, below,align = center] (f) at (a.south) {\large Whole-Body Planner};

\node [solid, draw = black,fill = black!50, minimum height = 6 cm, minimum width = 9.75 cm,text width = 3.375 cm,rounded corners=0.125 cm,right, yshift = -6.5 cm] (g) at (a.west) {};
\node[solid, below,rotate = 90,white] at (g.west) {\large Hierarchical QP};
\node [solid, draw = none,fill = none, minimum height = 0.5 cm, minimum width = 5.5 cm,text width = 5 cm,rounded corners=0.125 cm, below, align = center] (h) at (g.south) {\large Whole-Body Controller};

\node [solid, align = center, fill = red!50, minimum height = 1 cm, text width = 3.5 cm, minimum width = 4 cm,rounded corners=0.125 cm, yshift= -0.375 cm,xshift = -2.25cm, below] (i) at (g.north) {Equations of Motion};
\node [solid, align = center, fill = red!50,minimum height = 1 cm, text width = 3.5 cm, minimum width = 4 cm,rounded corners=0.125 cm, yshift= -0.375 cm, below] (j) at (i.south) {Zero Acceleration at Contact Feet};
\node [solid, align = center, fill = red!50, minimum height = 1 cm, text width = 3.5 cm, minimum width = 4 cm,rounded corners=0.125 cm, yshift= -0.375 cm, below] (k) at (j.south) {Joint Torque Limits};
\node [solid, align = center, fill = red!50,minimum height = 1 cm, text width = 3.5 cm, minimum width = 4 cm,rounded corners=0.125 cm, yshift= -0.375 cm, below] (l) at (k.south) {Friction Pyramid \\Constraints};

\node [solid, align = center, fill = white, minimum height = 1 cm, text width = 3.5 cm, minimum width = 4 cm,rounded corners=0.125 cm, yshift= -0.375 cm,xshift = 2.25cm, below] (m) at (g.north) {Base Motion Tracking};
\node [solid, align = center, fill = white, minimum height =1 cm, text width = 3.5 cm, minimum width = 4 cm,rounded corners=0.125 cm, yshift= -0.375 cm, below] (n) at (m.south) {Swing Feet \\ Trajectory Tracking};
\node [solid, align = center, fill = white, minimum height = 1 cm, text width = 3.5 cm, minimum width = 4 cm,rounded corners=0.125 cm, yshift= -0.375 cm, below] (o) at (n.south) {Arm Joint \\ Motion Tracking};
\node [solid, align = center, fill = white, minimum height =1 cm, text width = 3.5 cm, minimum width = 4 cm,rounded corners=0.125 cm, yshift= -0.375 cm, below] (p) at (o.south) {Arm Contact Force Tracking};

\draw[solid, thick,-stealth, rounded corners=0 cm] (e.south) |- (m) node[pos = 0.5] (a1){} node[pos = 0.7,above ] (bc) {$\bm \Phi^*_{IB},\bm \omega^*_{IB},  \dot{\bm \omega}^*_{IB}$};
\node[solid, above] at (bc.north) {$\bm r^*_{IB},\bm v^*_{IB},  \dot{\bm v}^*_{IB}$};
\draw[solid, thick,-stealth,rounded corners=0 cm] (a1.center) |- (n)   node[pos = 0.5] (a2){}  node[pos = 0.7,above ] {$\bm r^*_{c_i}, \dot {\bm r}^*_{c_i}, \ddot{\bm r}^*_{c_i}$};
\draw[solid, thick,-stealth,rounded corners=0 cm] (a2.center) |- (o)  node[pos = 0.5] (a3){}  node[pos = 0.7,above ] {$\bm q^*_{b}, \dot{\bm q}^*_{b}, \ddot{\bm q}^*_{b}$};
\draw[solid, thick,-stealth,rounded corners=0 cm] (a3.center) |- (p)   node[pos = 0.5] (a4){}  node[pos = 0.7,above ] {$\bm f_{c_5}$};
\end{tikzpicture}}};
\node [draw = black,fill = blue!30, minimum height = 1 cm, minimum width = 2.25 cm,text width = 2.25 cm,rounded corners=0.125 cm, yshift = -6.5 cm, align = center, xshift = 1cm ] (B) at (A.west) {High-Level \\ Controller};
\node [draw = black,fill = blue!30, minimum height = 1 cm, minimum width = 2.25 cm,text width = 2.25 cm,rounded corners=0.125 cm,xshift =1cm, align = center, right, yshift = -1 cm] (C) at (B.east) {Low-Level \\ Module};
\node [ draw = black,fill = blue!30, minimum height = 1 cm, minimum width = 2.25 cm,text width = 2.25 cm,rounded corners=0.125 cm,xshift =1 cm, align = center, right] (D) at (C.east) {Joint \\ Controller};
\node [ draw = black,fill = none, minimum height = 2 cm, minimum width = 2.25 cm,text width = 2.25 cm,rounded corners=0.125 cm,xshift =1cm, align = center, right] (E) at (D.east) {\includegraphics[scale=0.1]{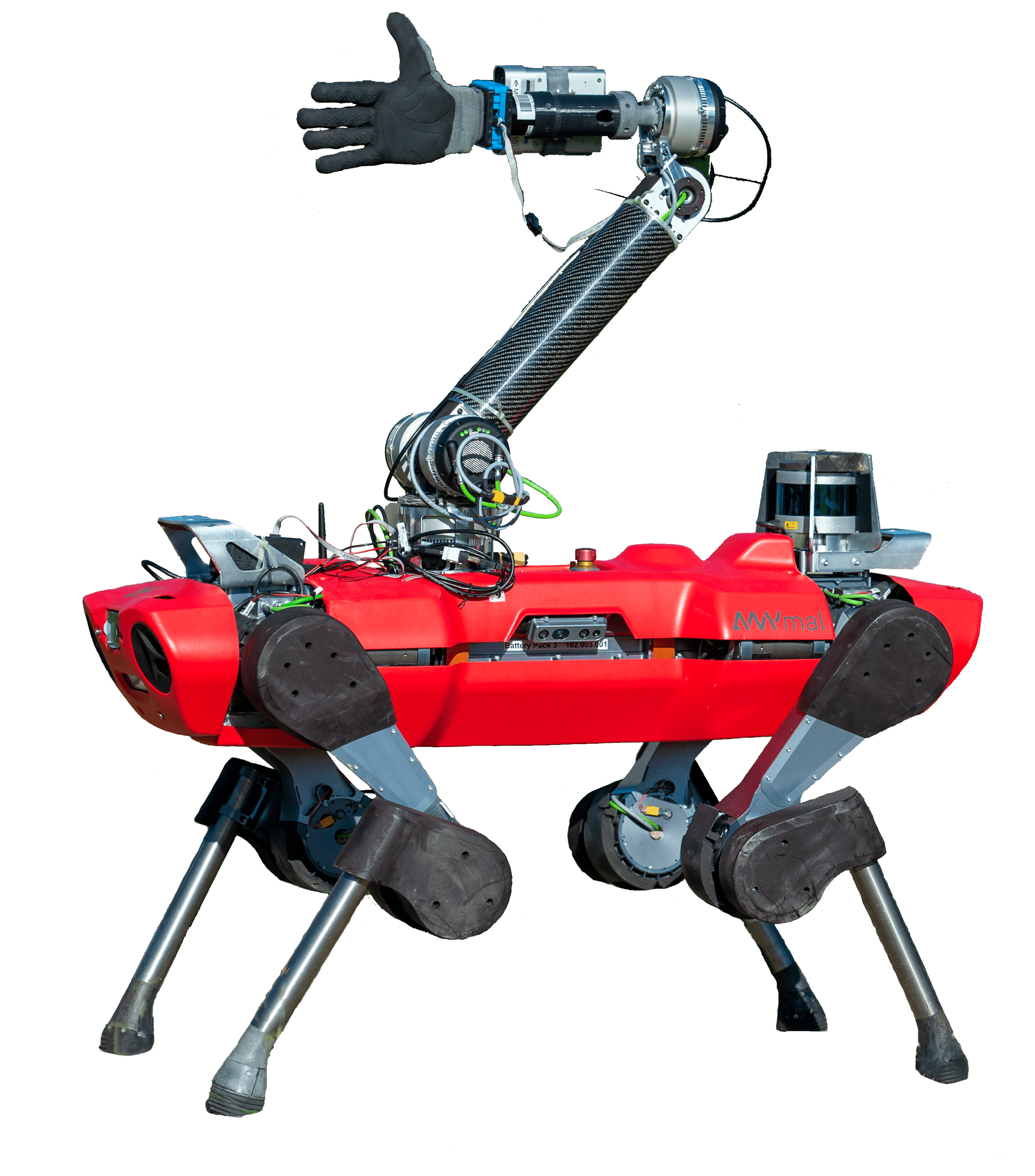}};
\node [ draw = black,fill = blue!30, minimum height = 1 cm, minimum width = 2.25 cm,text width = 2.25 cm,rounded corners=0.125 cm,yshift =- 1cm, align = center, below] (F) at (B.south) {State\\ Estimator};
\node[yshift=0.25cm] (BB) at (C.west) {};
\node[yshift=-0.25cm] (CC) at (C.west) {};
\node[yshift=0.25cm] (DD) at (C.east) {};
\node[yshift=-0.25cm] (EEE) at (C.east) {};
\node[yshift=0.25cm] (FF) at (D.west) {};
\node[yshift=-0.25cm] (GG) at (D.west) {};

\draw[thick,-stealth] (B.east)-- ++(0.5cm,0cm)node[pos= 0.25] (yyy) {} --++(0cm,-0.75cm) -- (BB.center) ;
\draw[thick,stealth-] (F.east)-- ++(0.5cm,0cm)node [pos= 0.25, below] (xxx) {} --++(0cm,0.75cm) -- (CC.center);
\node[circle, draw= black, inner sep=0pt,minimum size=2.75cm] (C) at (B) {};
\node[xshift= 1.5cm] (EE) at (A.south west) {};
\draw[thick,-stealth] (F.west)-- ++(-0.5cm,0cm)  |- (B.west) node[pos= 0.25, left,text width = 0.75 cm, align=center] {$\bm q_b$, \\ $ \bm \omega_{IB}$, \\ $\bm q_j$, \\ $\dot{\bm q}_j$};
\node[anchor = north west] at (xxx) {$\bm q_j$, $\dot{\bm q}_j$};
\node[anchor = south west] at (yyy) {$\bm \tau^*_j$, $\bm q^*_j$, $\dot{\bm q}^*_j$};
\draw[] (C.90) -- (EE.center);
\draw[thick,-stealth](DD.center) -- (FF.center) node[pos= 0.5, above] {};
\draw[thick,stealth-](EEE.center) -- (GG.center)node[pos= 0.5, below] {};
\draw[thick,-stealth](D.east) -- (E.west)node[pos= 0.5, above] {$\bm \tau_a$};
\end{tikzpicture}}
    \caption{A schematic diagram depicting the various components of the full control architecture. The high-level control module is highlighted and detailed in the top diagram. This consists of the whole-body planner interacting with the WBC whose high-priority tasks are indicated in red.}
    \label{fig:MPC-WBC}
    \vspace*{0mm}
\end{figure}

The optimal reference plans for the base and limbs are tracked by a whole-body controller (WBC) that tries to fulfill a set of prioritized tasks. These tasks are formulated in the form of a hierarchical quadratic program (QP) that optimizes for the generalized accelerations and contact forces. Joint torques are then retrieved by inverting the desired dynamics. The objectives with their corresponding priorities are represented in the high-level controller diagram of~\cref{fig:MPC-WBC}. For a detailed discussion on how the QP structure is set up and solved, the reader is referred to the relevant papers \cite{WBC1,WBC2}. 

It is worth noting that we do not aim to directly track the optimal ground reaction forces; instead, we capture their influence by tracking the reference motion they induce on the base. The reason for that is related to the fact that the WBC imposes stricter conditions on physical correctness than the planner does, as it relies on a more realistic model. Consequently, the forces are adjusted if they violate any higher priority objectives, so by having a base motion task we direct the solver to redistribute the forces and accelerations in such a way that would, at the very least, keep the robot balanced. On the other hand, arm contact forces outputted by the MPC planner are treated differently and sent as direct references to the QP. This is because the WBC has no knowledge of the manipulated object's dynamics. Therefore, it would not be possible to define a prioritized task on the level of the object's motion. 

It is also important to highlight the conversions needed to couple the MPC output solutions with the WBC tracking tasks. To begin with, the arm's reference joint positions, joint velocities, and contact forces are directly accessible from $\bm x^*$ and $\bm u^*$, whereas the joint accelerations $\ddot{\bm q}_j$ are approximated by finite differences. Swing feet trajectories are obtained from basic kinematic transformations, and by using the expression below for task-space accelerations 
\begin{equation}
    \ddot{\bm r}_{c_i} = \bm J_{c_i} \ddot{\bm q}_j + \dot{\bm J}_{c_i} \dot{\bm q}_j.
\end{equation}
Similarly, the desired base pose $\bm q^*_b$ is part of the MPC optimal state, while its linear and angular velocities can be extracted from~\cref{eq:BaseDerivative} with a simple mapping
\begin{equation} \label{eq:baseVel}
    \begin{bmatrix} \bm v_{IB} \\ \bm \omega_{IB} \end{bmatrix} = \begin{bmatrix} \mathbb{I}_{3\times3} & \bm 0_{3\times3} \\
                    \bm 0_{3\times3} & \bm T(\bm \Phi^{zyx}_{IB})
    \end{bmatrix} \dot{\bm q}_b,
\end{equation}
where $\bm T \in \mathbb{R}^{3\times3}$ maps derivatives of ZYX-Euler angles to angular velocities. As for the feedforward base accelerations, we start by deriving the expression for $\ddot{\bm q}_b$ from~\cref{eq:centroidaldynamics2} by taking the time-derivative on both sides and rearranging terms:
\begin{equation} \label{eq:ddotQ}
    \ddot{\bm q}_b = \bm A_b^{-1} \left(\dot{\bm h}_{com} - \dot{\bm A} \dot{\bm q} - \bm A_j \ddot{\bm q}_j \right)
\end{equation}
where $\dot{\bm h}_{com}$ is given by~\cref{eq:CentroidalDynamics} and $\dot{\bm A} \dot{\bm q}$ is retrieved from the Recursive Newton-Euler Algorithm (RNEA) by setting the joint accelerations to zero and transforming the resulting base wrench into a centroidal wrench. Finally, the computation of the accelerations trivially follows from~\cref{eq:baseVel,eq:ddotQ} 
\begin{equation}
    \begin{bmatrix} \dot{\bm v}_{IB} \\ \dot{\bm \omega}_{IB} \end{bmatrix} = \begin{bmatrix} \mathbb{I}_{3\times3} & \bm 0_{3\times3} \\
                    \bm 0_{3\times3} & \bm T
    \end{bmatrix} \ddot{\bm q}_b + 
    \begin{bmatrix} \bm 0_{3\times3} & \bm 0_{3\times3} \\
                    \bm 0_{3\times3} & \dot{\bm T}
    \end{bmatrix} \dot{\bm q}_b.
\end{equation} 
\section{SYSTEM DESCRIPTION}
The framework described in~\cref{Formulation} is robot-agnostic and is thus capable of encompassing, with the possibility of minor modifications, a wide variety of multi-limbed systems designed for locomotion and manipulation purposes. In this work, we demonstrate the validity of our approach through a set of experiments performed on a quadrupedal mobile manipulator consisting of the ANYmal C platform equipped with the DynaArm, a custom-made 4-DoF robotic arm. The latter is a newly developed torque-controllable manipulator, which comprises four powerful actuators that allow for highly dynamic maneuvers and a high payload capability of 7 kg. The distinguishing feature of the DynaArm is that the elbow flexion joint is driven by an actuator that is situated at the shoulder
through a belt transmission mechanism. This has the advantage of reducing the torque load on the shoulder flexion joint, since the reaction torque caused by the elbow drive is now directly transmitted to the arm's base. 

The full control architecture is depicted in the schematic diagram of~\cref{fig:MPC-WBC}. Apart from the joint controller, all of the modules run on the robot's onboard computer (Intel Core i7-8850H CPU@4GHz hexacore processor). With a time horizon of ${T = 1 \ \text{s}}$, the MPC loop computes feedforward trajectories at an average update rate of 70 Hz (in free-motion, when no object state is augmented to the MPC formulation). Both the whole body planner and controller obtain their feedback from a state-estimator that fuses encoder readings and IMU measurements to estimate the base pose. The WBC along with the state-estimator constitute the main control loop which runs at 400 Hz. Finally, the low-level module communicates back-and-forth data with the drive controller, where the actuator torque commands are generated at a frequency of 2.5 kHz:
\begin{equation}
    \bm \tau_a = \bm \tau^*_j + \bm K_p (\bm q^*_{j} - \bm q_j) + \bm K_d (\dot{\bm q}^*_{j} - \dot{\bm q}_j).
\end{equation}
\section{EXPERIMENTAL RESULTS} \label{Experiments}
We perform various experiments both in simulation (visualizations of the centroidal dynamics under the MPC policy) and on real hardware. The experiments are divided into two main categories, ones done in free-motion and others involving object-manipulation tasks. All of the examples discussed in this paper are also included in the supplementary video submission \href{https://youtu.be/uT4ypNDzUvI}{\textcolor{blue}{(Link)}}. We start by presenting the generic cost function used in the upcoming test cases: 
\begin{align}
L(\bm x, \bm u, t)
&= \, \alpha_1 \cdot \left(||\bm r_{IE} - \bm r_{IE}^{ref}||^2_{\bm Q_{ee_p}} + ||\bm \zeta_{IE}||^2_{\bm Q_{ee_o}}\right) \notag
\\ 
&+ \alpha_2 \cdot ||\bm x_r - \bm x_r^{ref}||^2_{\bm Q_r} + \alpha_3 \cdot  ||\bm x_o - \bm x_o^{ref}||^2_{\bm Q_o} \notag
\\ 
&+ ||\bm u - \bm u^{ref}||^2_{\bm R}
\end{align}
where $\bm R$ is positive definite and ${\bm Q_r, \ \bm Q_o, \ \bm Q_{ee_p}, \ \bm Q_{ee_o}}$ are positive semi-definite weighting matrices. The vectors ${\bm r_{IE}, \bm \zeta_{IE} \in \mathbb{R}^3}$ correspond to the arm's end-effector position and the orientation error (represented with exponential coordinates), respectively. Whereas the parameters ${\alpha_1, \alpha_2, \alpha_3 \in \{0,1\}}$ are used to switch between different objectives depending on the nature of the task.  
\subsection{Free-Motion}
\begin{figure}[t]
    \centering
    \begin{subfigure}[t]{0.5\textwidth}
        \centering
        \scalebox{0.75}{
        \begin{tikzpicture}
\node (A) at (0,0) {\includegraphics[trim={30cm 12cm 30cm 0cm},clip,scale=0.05]{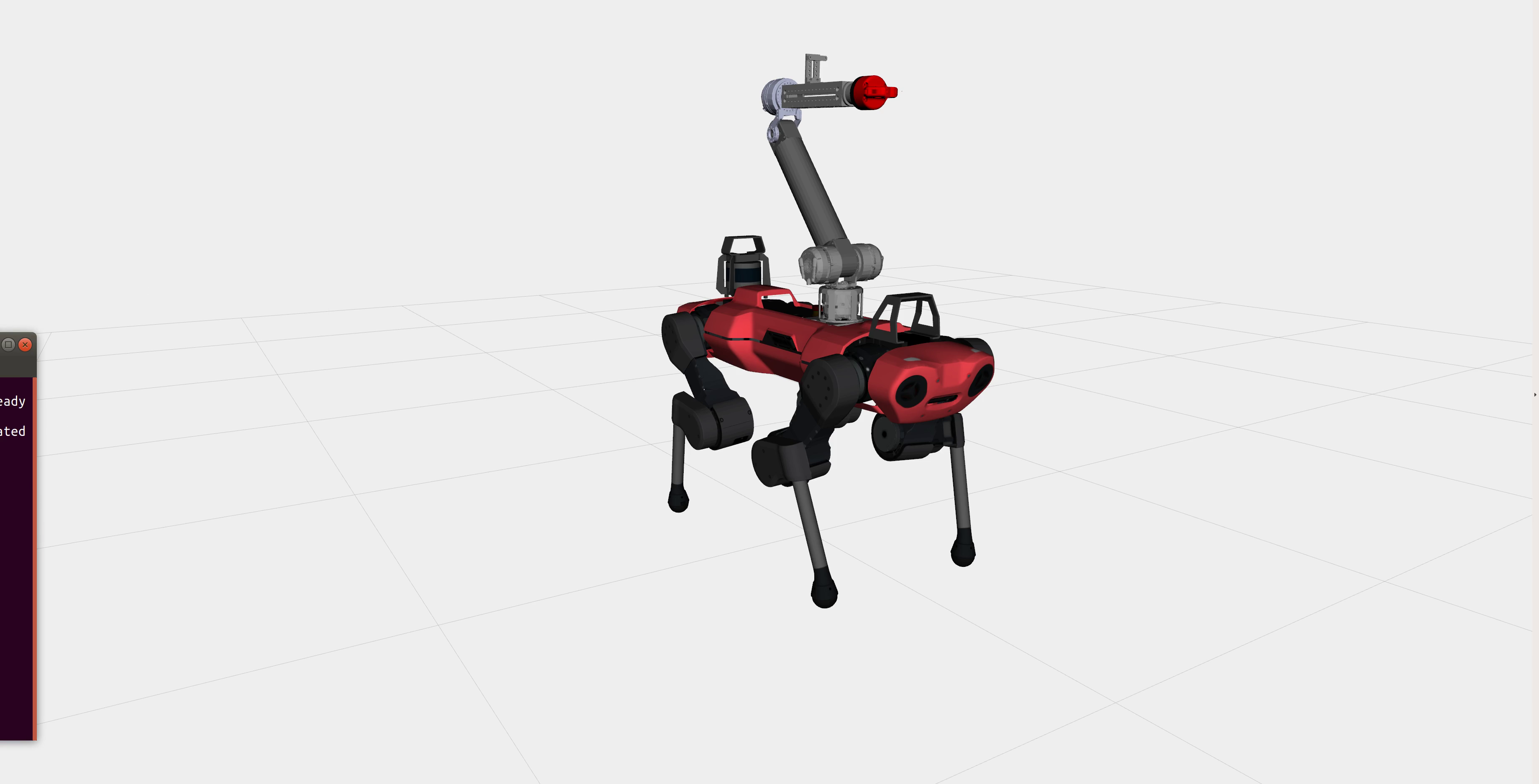}};
\node[right] (B) at (A.east) {\includegraphics[trim={30cm 12cm 30cm 0cm},clip,scale=0.05]{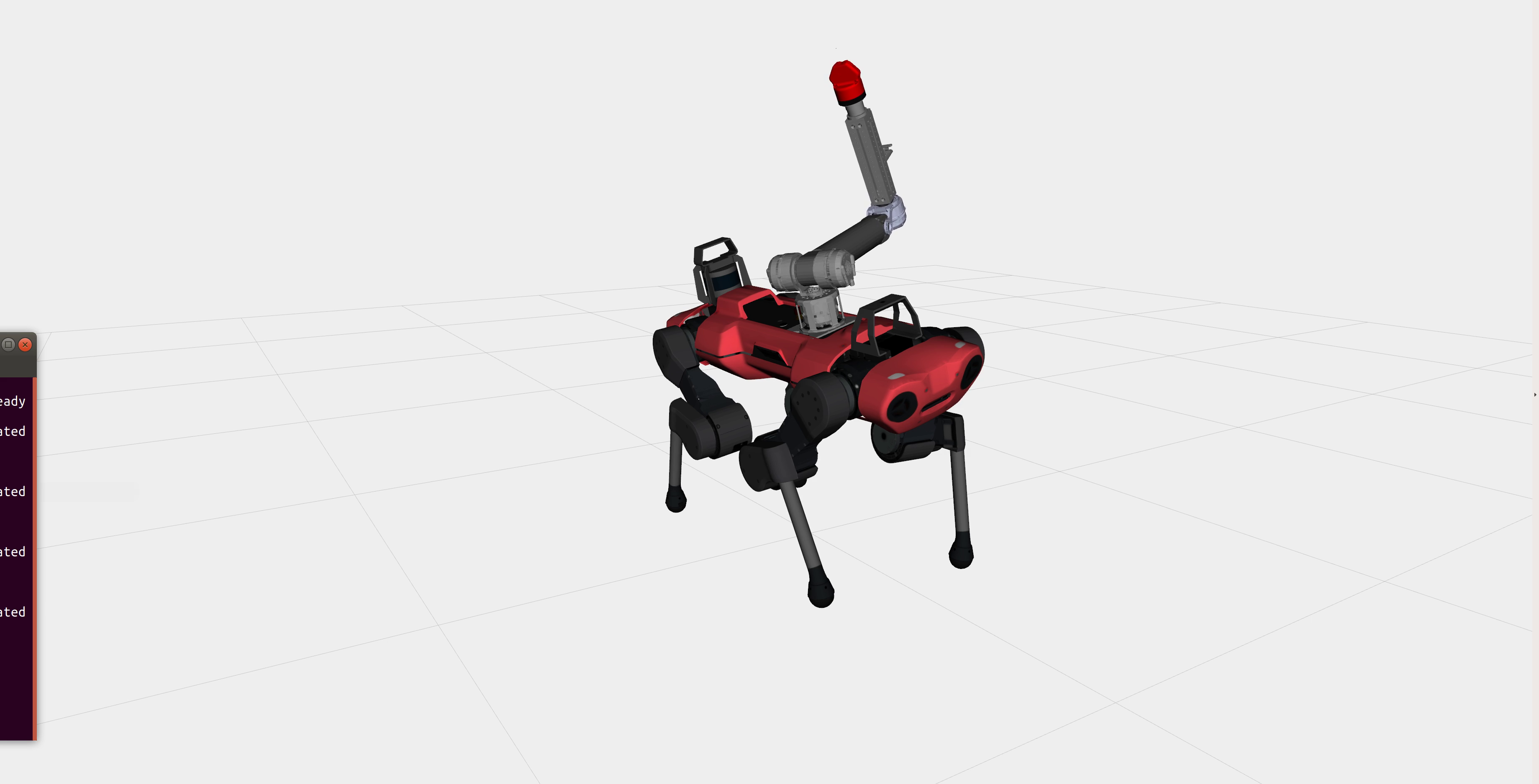}};
\node[right] (C) at (B.east) {\includegraphics[trim={30cm 12cm 30cm 0cm},clip,scale=0.05]{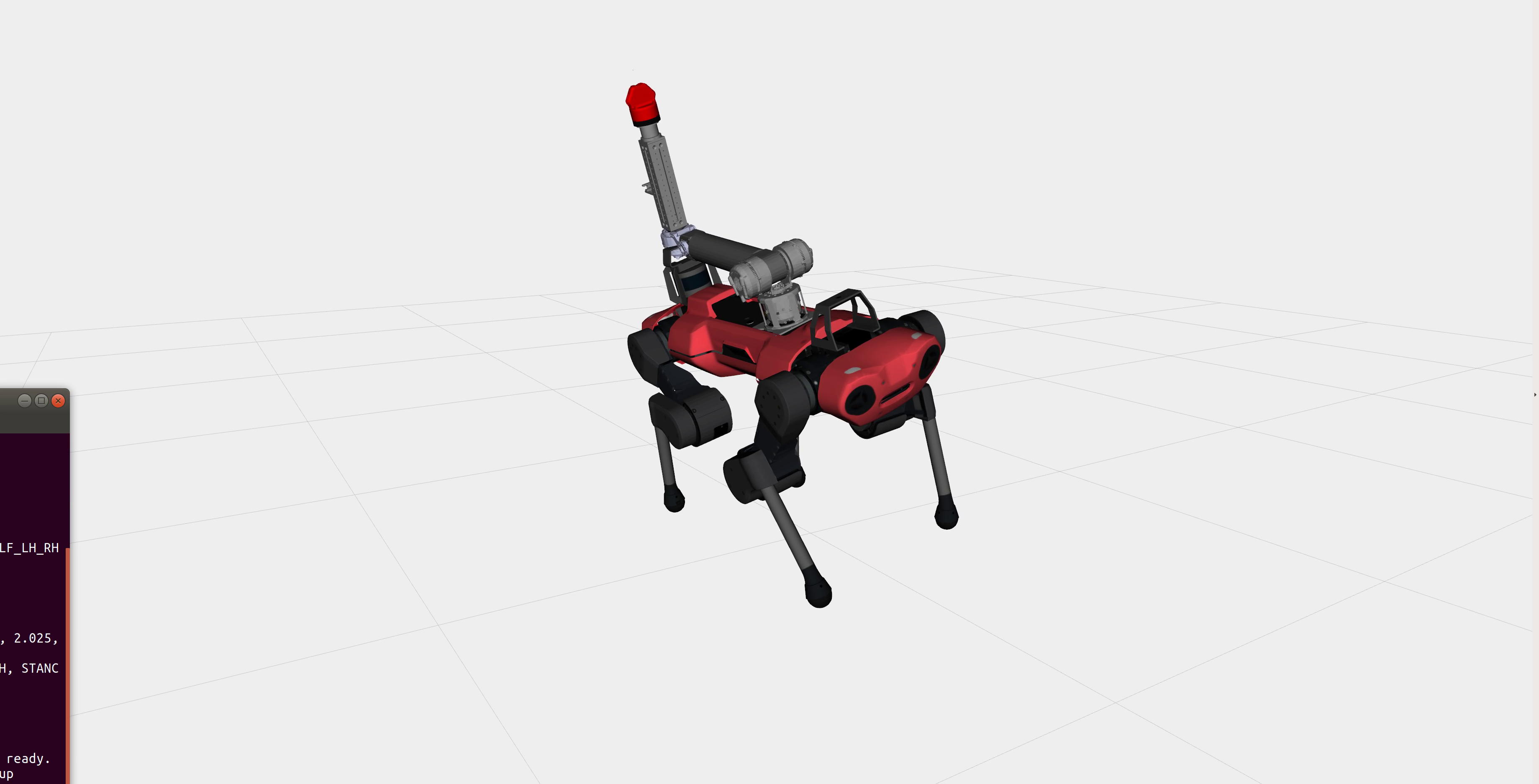}};
\node(AA) at (A.north west) {};
\node [ draw = black, minimum height = 0.3 cm, minimum width = 0.4cm, xshift = 0.675cm,yshift = -0.25cm, anchor = north east] at (AA) {\small 1};
\node(AA) at (B.north west) {};
\node [ draw = black, minimum height = 0.3 cm, minimum width = 0.4cm, xshift = 0.675cm,yshift = -0.25cm, anchor = north east] at (AA) {\small 2};
\node(AA) at (C.north west) {};
\node [ draw = black, minimum height = 0.3 cm, minimum width = 0.4cm, xshift = 0.675cm,yshift = -0.25cm, anchor = north east] at (AA) {\small 3};
\end{tikzpicture}}
        \caption{\textbf{1.} Robot starts in a nominal stance mode. \textbf{2.} Base is commanded to roll with a 30 degree angle. \textbf{3.} Left fore leg is lifted.}
        \label{fig:Tail_a}
    \end{subfigure}
    
    \begin{subfigure}[t]{0.5\textwidth}
        \centering
                \scalebox{0.75}{
       \begin{tikzpicture}
\node (A) at (0,0) {\includegraphics[trim={30cm 12cm 30cm 0cm},clip,scale=0.05]{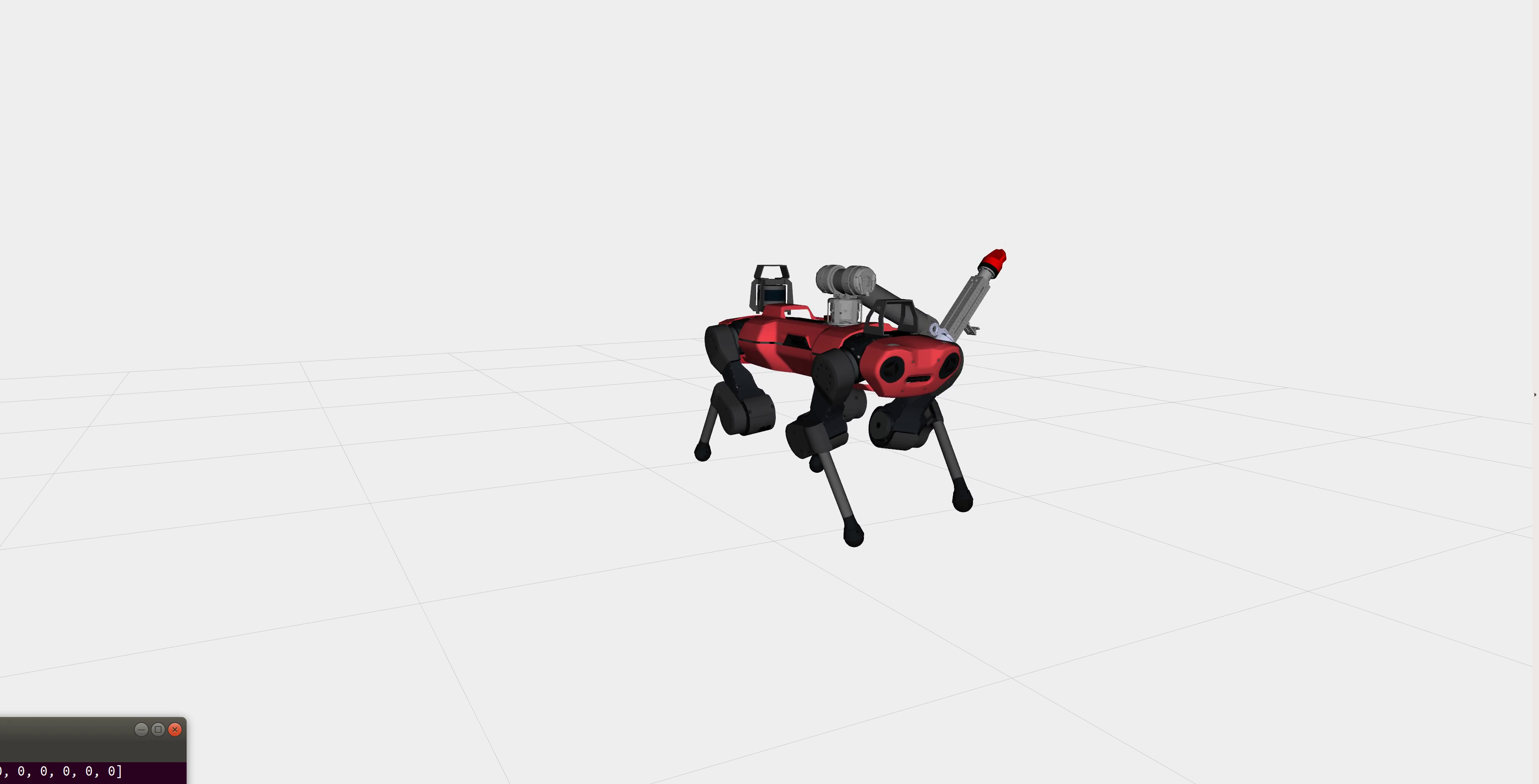}};
\node[right] (B) at (A.east) {\includegraphics[trim={30cm 12cm 30cm 0cm},clip,scale=0.05]{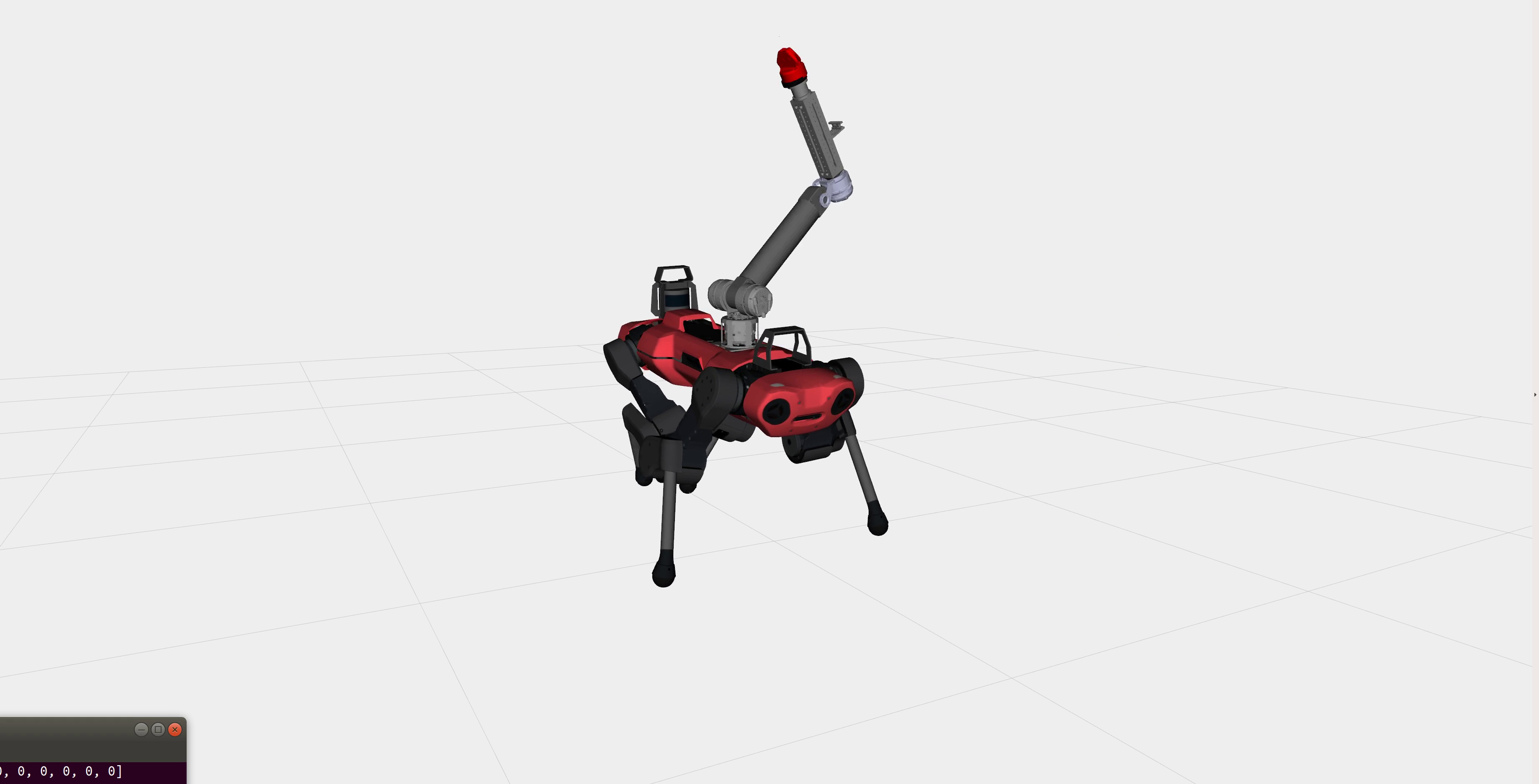}};
\node[right] (C) at (B.east) {\includegraphics[trim={30cm 12cm 30cm 0cm},clip,scale=0.05]{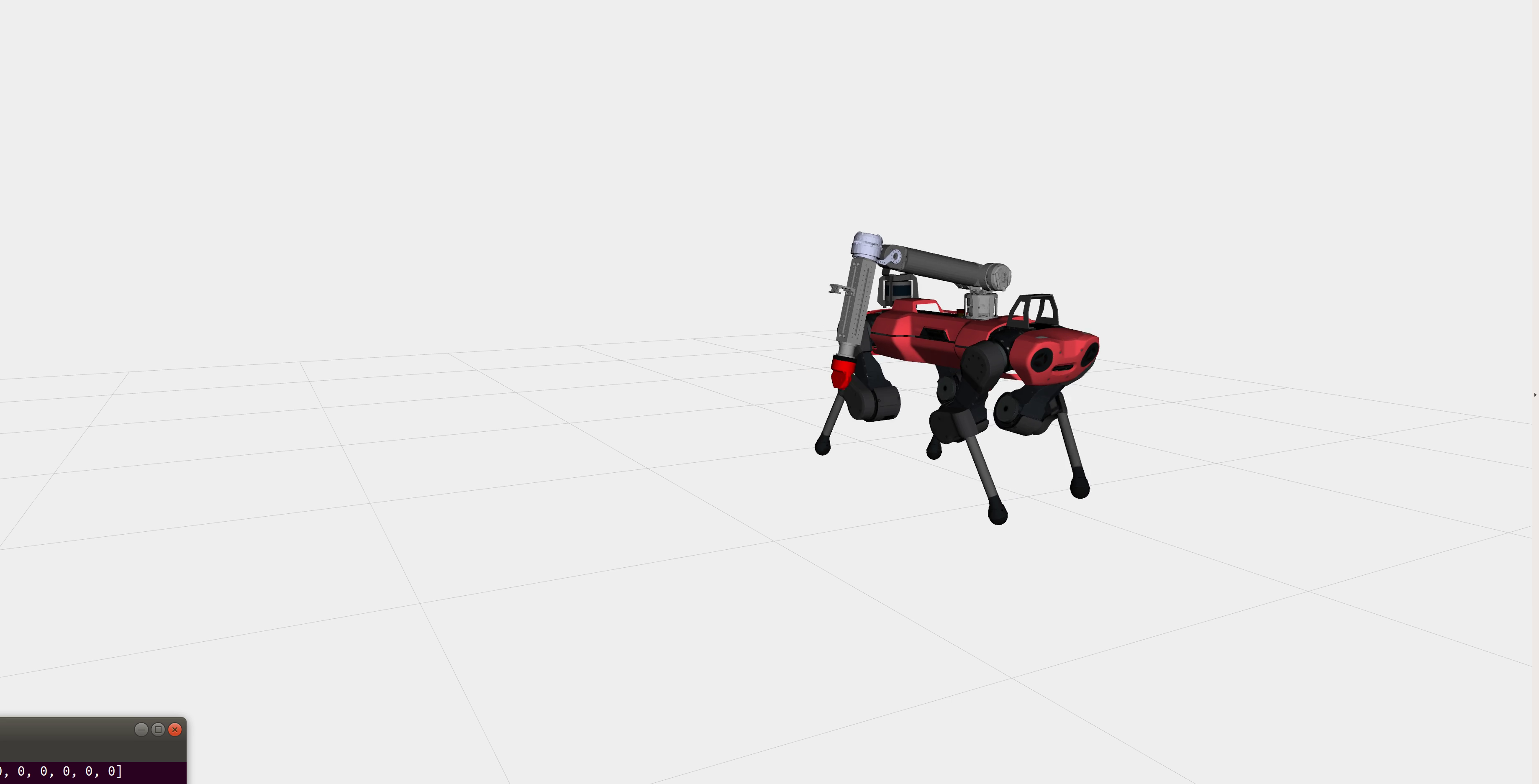}};
\draw[very thick,stealth-] (A.center) ++ (0cm,-1cm) ++ (-1cm,0cm) -- ++(1cm,0cm);
\draw[very thick,-stealth] (C.center) ++ (0cm,-1cm) ++ (-1cm,0cm) -- ++(1cm,0cm);
\node(AA) at (A.north west) {};
\node [ draw = black, minimum height = 0.3 cm, minimum width = 0.4cm, xshift = 0.675cm,yshift = -0.25cm, anchor = north east] at (AA) {\small 1};
\node(AA) at (B.north west) {};
\node [ draw = black, minimum height = 0.3 cm, minimum width = 0.4cm, xshift = 0.675cm,yshift = -0.25cm, anchor = north east] at (AA) {\small 2};
\node(AA) at (C.north west) {};
\node [ draw = black, minimum height = 0.3 cm, minimum width = 0.4cm, xshift = 0.675cm,yshift = -0.25cm, anchor = north east] at (AA) {\small 3};
\end{tikzpicture}}
        \caption{\textbf{1.} Robot is trotting laterally in one direction. \textbf{2.} Robot is commanded to switch directions. \textbf{3.} Robot is trotting laterally in the other direction.}
        \label{fig:Tail_b}
    \end{subfigure}
    \caption{Snapshots demonstrating the use of the arm as a ``tail" that aids in balancing. The sequences correspond to (a) static and (b) dynamic balancing scenarios.}
    \label{fig:tail}
\end{figure}
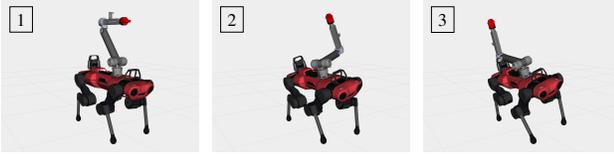
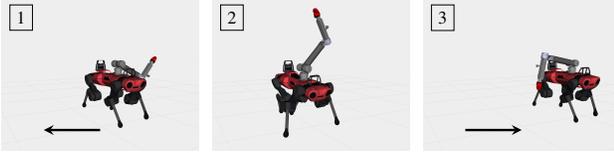

\begin{figure}[t]
    \centering
    \includegraphics[scale=0.36]{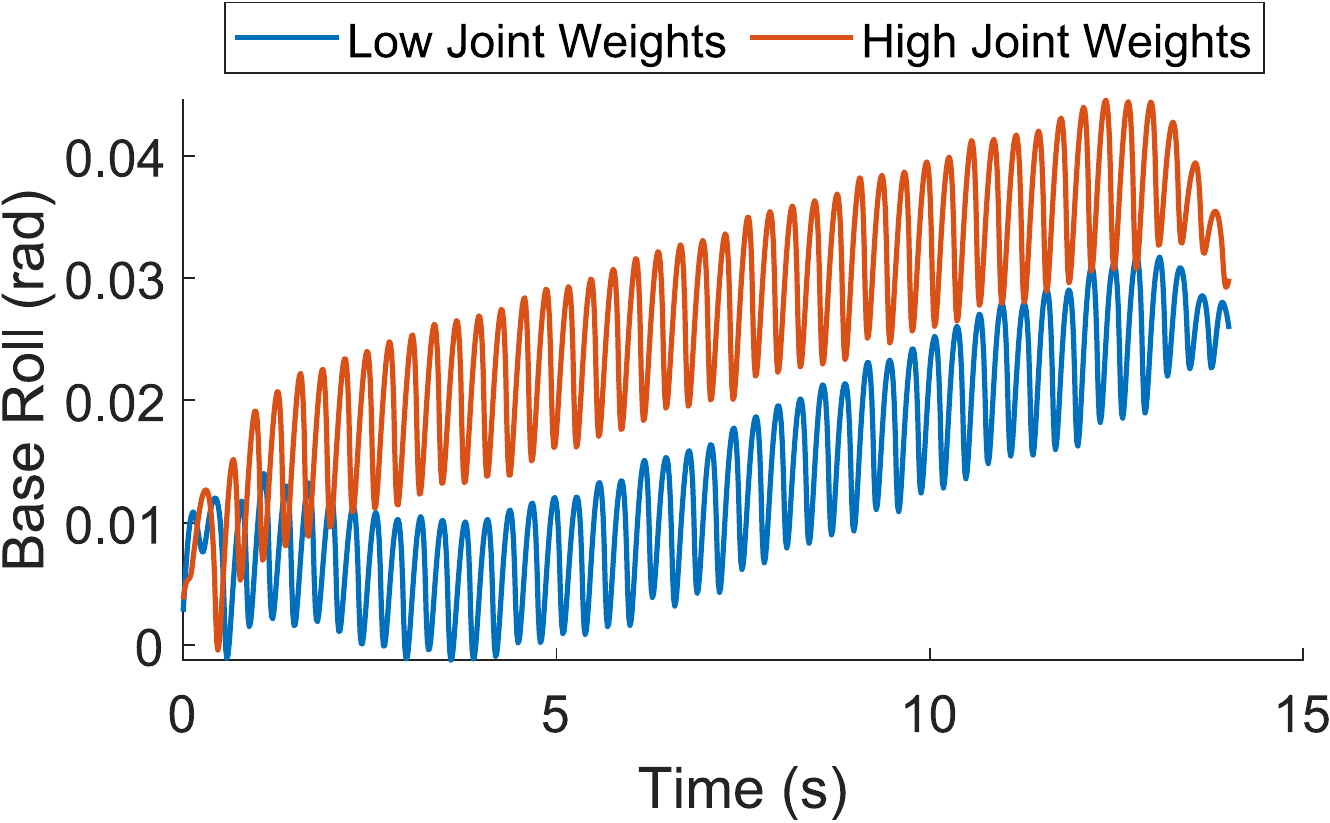}
    \caption{Plots showing the base's roll during a lateral trot in two scenarios: The arm is treated as a rigidly attached mass (high joint weights) -- The arm acts as a balancing ``tail" (low joint weights).}
    \label{fig:baseRoll}
        \vspace{0mm}
\end{figure}

In this section, we set $\alpha_3 = 0$ for all experiments. The first set of examples focuses on commanding the base motion which is already part of the state; thus we set $\alpha_1 = 0$ and $\alpha_2 = 1$ initially. We start with a simple test on the real system, where we demonstrate two different dynamic gaits, namely a trot and flying trot, whose mode sequences are depicted in the video. A high penalty is imposed on the arm joint positions and velocities to keep it at a nominal configuration, thereby leading the planner to treat it as a lumped mass with respect to the base. It is clear that in such cases the arm does not play any role with regards to balancing.
Alternatively, reducing the joint weights allows the planner to exploit the base-limb coupling by using the arm as a ``tail" that contributes in balancing the robot. This behavior is showcased  in~\cref{fig:Tail_a} for a static balancing scenario, where we command the base with a 30 degree roll-angle while in stance mode, then we lift the left-fore leg. As a result, the arm moves so as to shift the whole-body center of mass in a direction that would redistribute the contact forces equally. Another scenario is shown in~\cref{fig:Tail_b}, where we command the robot to trot sideways with a relatively high velocity in both directions. As the robot accelerates in one direction, the arm swings in the opposing direction, and it does so by rotating about the x-axis of the base frame. Apart from force redistribution, this dynamic movement helps the base accelerate, while additionally counteracting the rolling centroidal angular-momentum induced by the lateral forces (as a consequence of~\cref{eq:CentroidalDynamics} and~\cref{eq:BaseDerivative}). This in turn helps regularize the base roll at the desired zero set-point. In fact, we compare the cases of a ``rigid" arm and a ``tail" arm for a reference lateral displacement, and plot the trajectories for the base roll in~\cref{fig:baseRoll}. We note a $66.8\%$ reduction in the $\mathcal{L}_2$-norm of the low-weight signal with respect to the high-weight signal. A similar behavior is also exhibited when commanding accelerations in the longitudinal directions where the arm's swinging motion helps regularize the base pitch. These examples are presented in simulation only to avoid potential collisions between the arm and base that could damage the platform.  
In the second set of experiments we focus on commanding the manipulator's end-effector. Accordingly, a task-space objective is assigned by setting $\alpha_1 = 1$ and $\alpha_2 = 1$, where the quadratic state-cost is used as a regularization term. 
To begin with, we carry out a task involving the manipulator reaching out to grasp an object or to place it. The perceived behavior is one where the torso clearly adapts its pose in coordination with the arm's movement (see attached video). This indeed highlights the importance of including the robot's full kinematic model in the planning phase.
By that we obtain motion plans that would optimally exploit the kinematic redundancy in our system to help achieve the end-effector tracking task. The degrees of freedom that are predominantly employed are determined by the relative weights in the matrix $\bm Q_r$. 
A second scenario consists of sending references to both the base and the gripper simultaneously. The gripper's position with respect to the inertial frame is fixed, while the base is commanded to trot back and forth. This is done while having the robot carry an unmodeled 2 kg load to illustrate the robustness of our control framework in its ability to overcome modeling errors.
\vspace{0mm}
\subsection{Object-Manipulation}
\begin{figure}[t]
\centering
    \begin{subfigure}[t]{0.25\textwidth}
        \centering
        \includegraphics[scale=0.3]{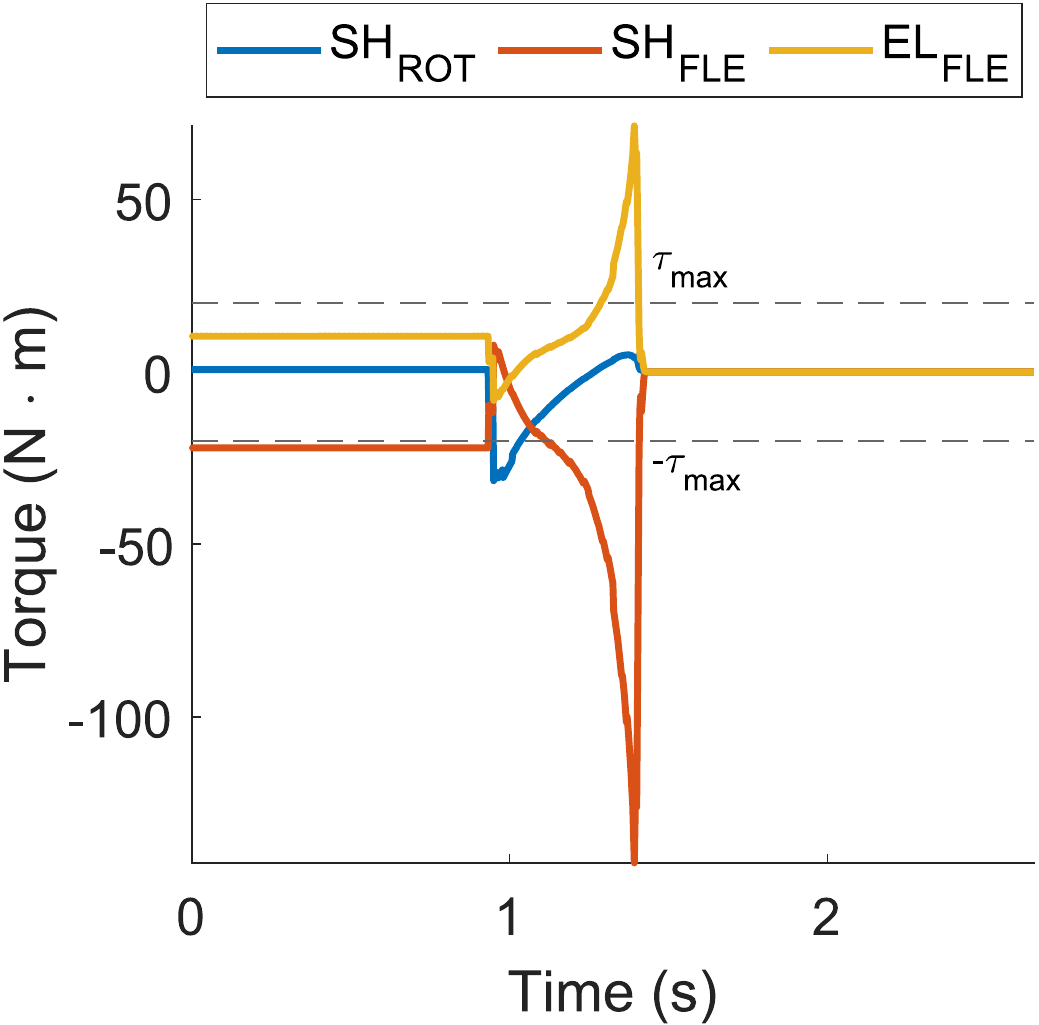}
    \end{subfigure}%
    \begin{subfigure}[t]{0.25\textwidth}
        \centering
        \includegraphics[scale=0.3]{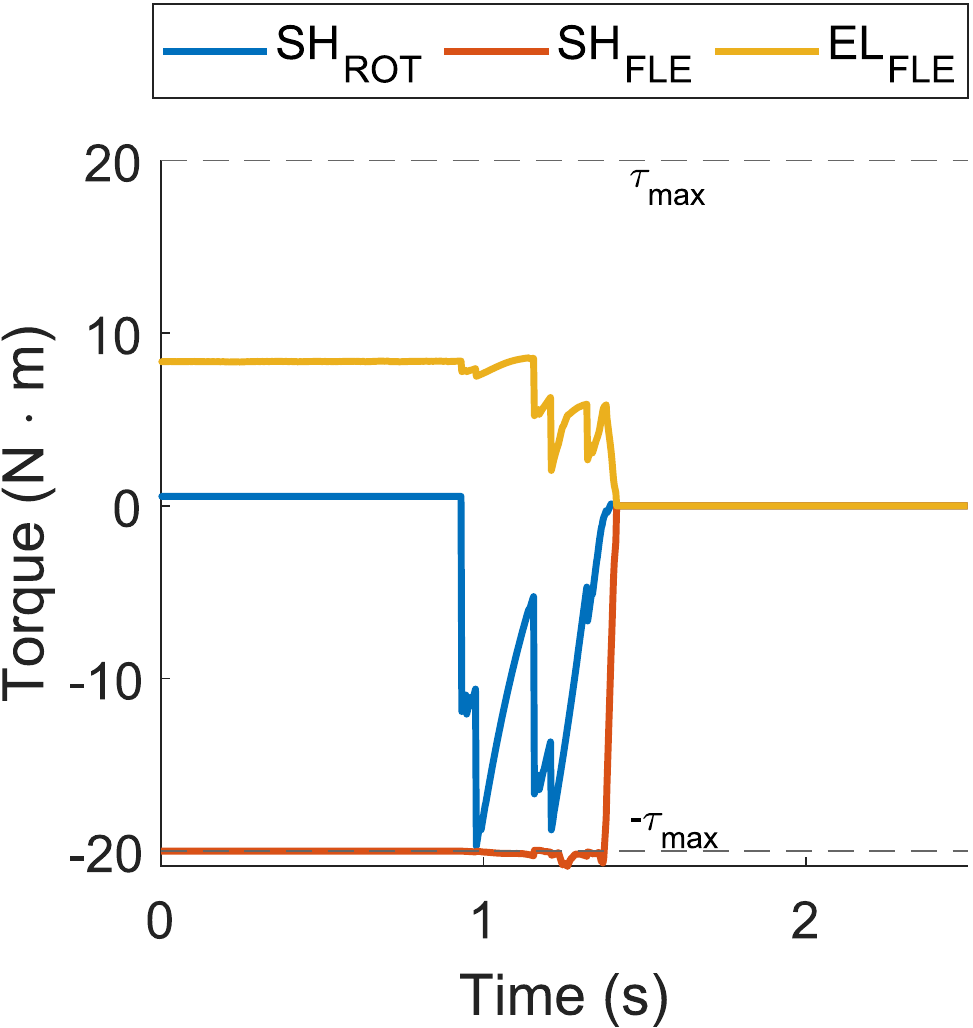}
    \end{subfigure}
    \caption{Plots showing the arm's joint torques for the dynamic throwing task without torque limits \textit{(left)} and with torque limits \textit{(right)}}
    \label{fig:torquePlots}
    \vspace{0mm}
\end{figure}

\begin{figure}[t]
    \centering
        \scalebox{0.75}{
        \begin{tikzpicture}
\node (A) at (0,0) {\includegraphics[trim={40cm 12cm 20cm 0cm},clip,scale=0.05]{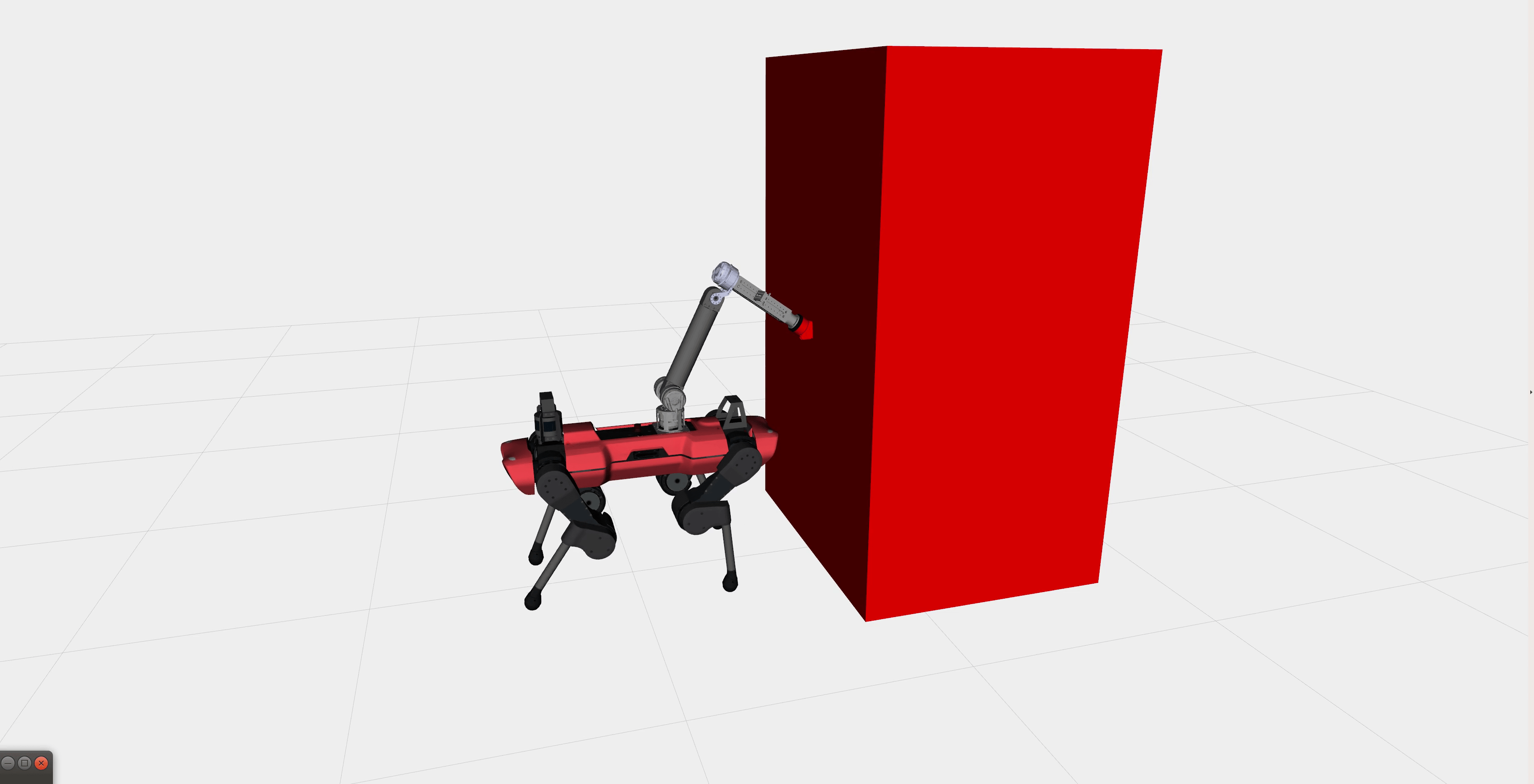}};
\node[right] (B) at (A.east) {\includegraphics[trim={40cm 12cm 20cm 0cm},clip,scale=0.05]{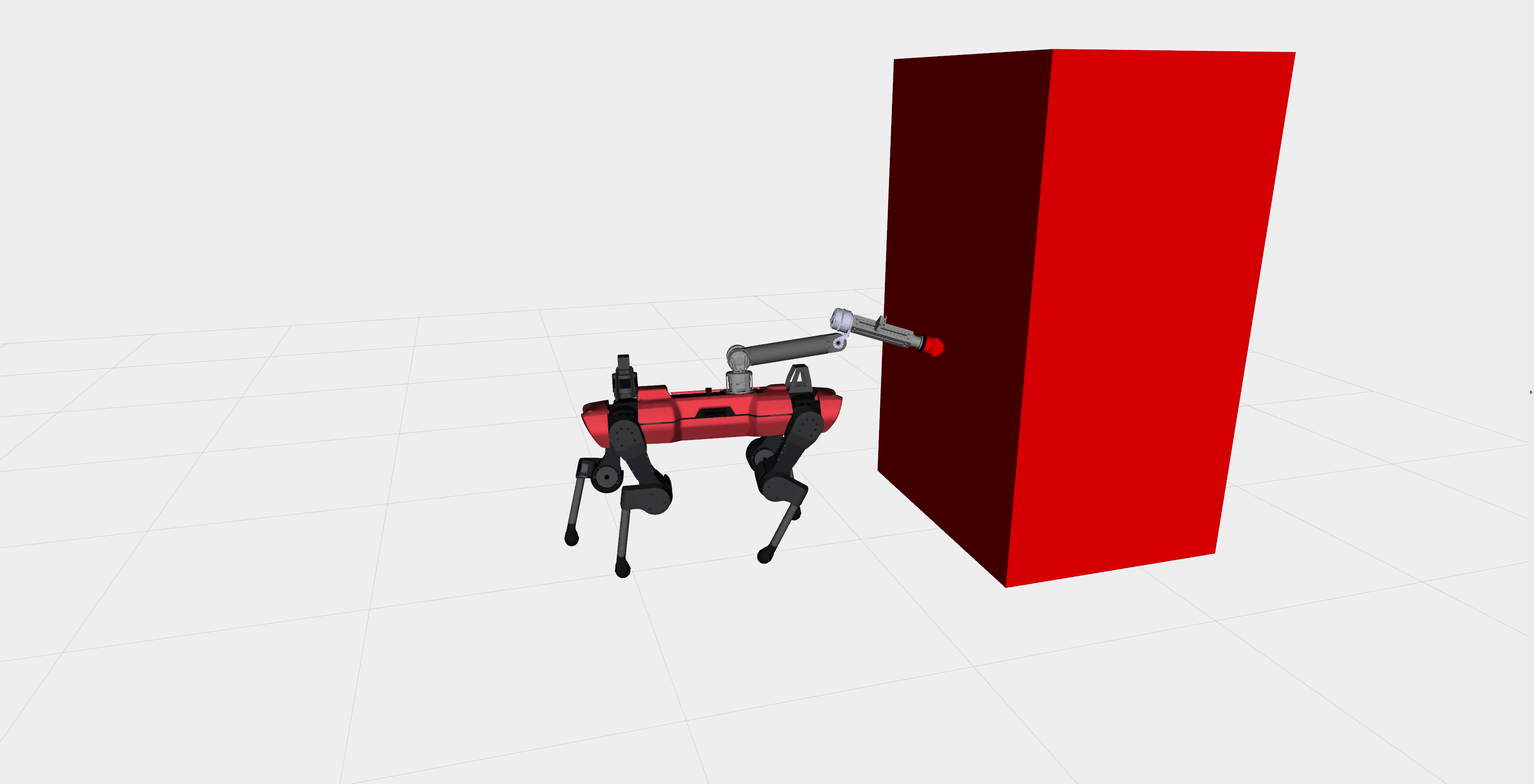}};
\node[right] (C) at (B.east) {\includegraphics[trim={40cm 12cm 20cm 0cm},clip,scale=0.05]{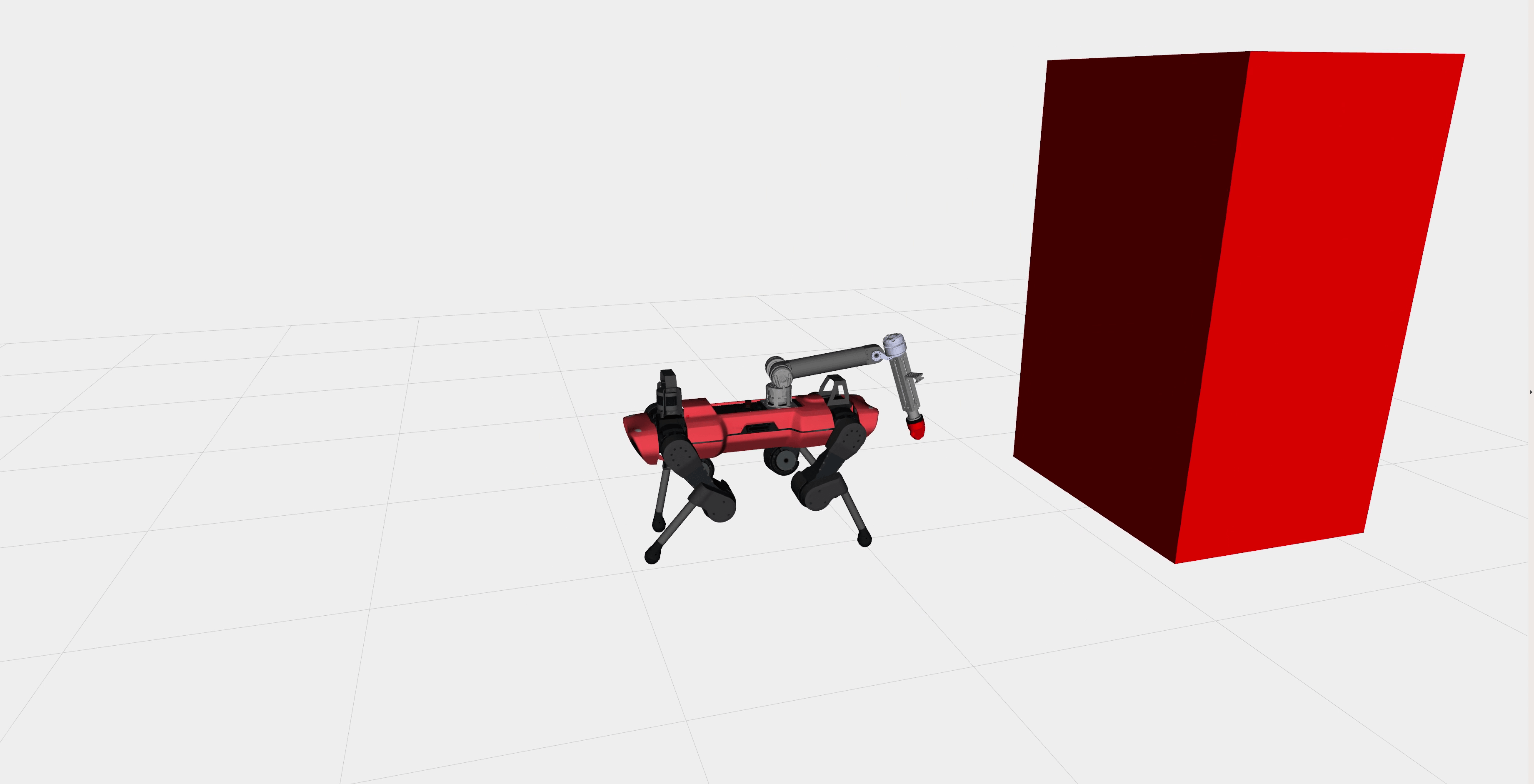}};
\node(AA) at (A.north west) {};
\node [ draw = black, minimum height = 0.3 cm, minimum width = 0.4cm, xshift = 0.675cm,yshift = -0.25cm, anchor = north east] at (AA) {\small 1};
\node(AA) at (B.north west) {};
\node [ draw = black, minimum height = 0.3 cm, minimum width = 0.4cm, xshift = 0.675cm,yshift = -0.25cm, anchor = north east] at (AA) {\small 2};
\node(AA) at (C.north west) {};
\node [ draw = black, minimum height = 0.3 cm, minimum width = 0.4cm, xshift = 0.675cm,yshift = -0.25cm, anchor = north east] at (AA) {\small 3};
\end{tikzpicture}}
        \caption{Snapshots of the robot dynamically pushing a 10 kg load towards a target position.}
        \label{fig:Pushing}
        \vspace{0mm}
\end{figure}
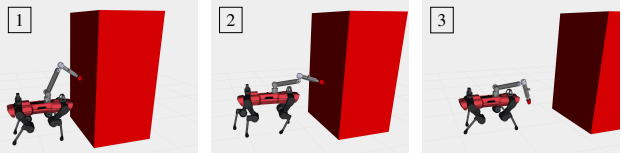

\begin{figure}[t]
    \centering
\begin{subfigure}[t]{0.5\textwidth}
        \centering
 \begin{tikzpicture}
    \node (A) at (0,0) {\includegraphics[scale=0.26]{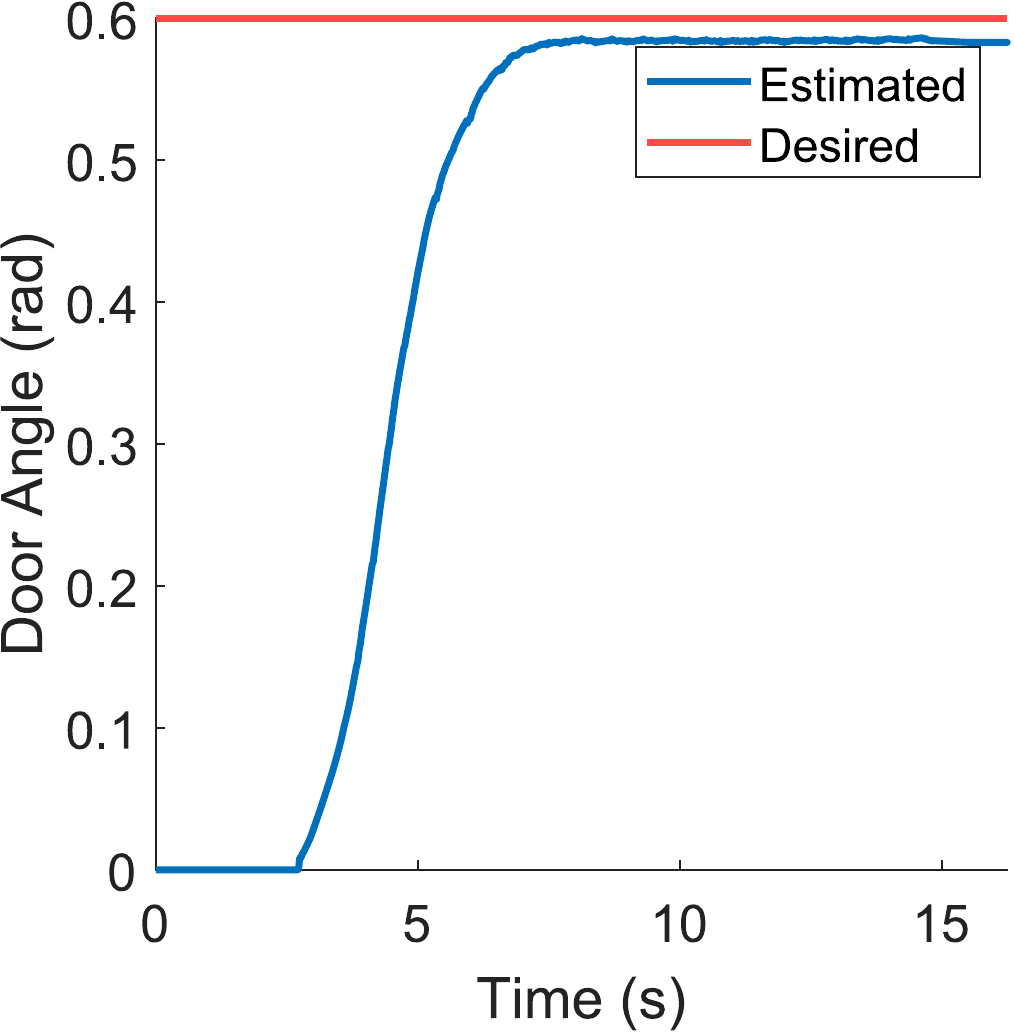}};
    \node[right] (B) at (A.east) {\includegraphics[scale=0.26]{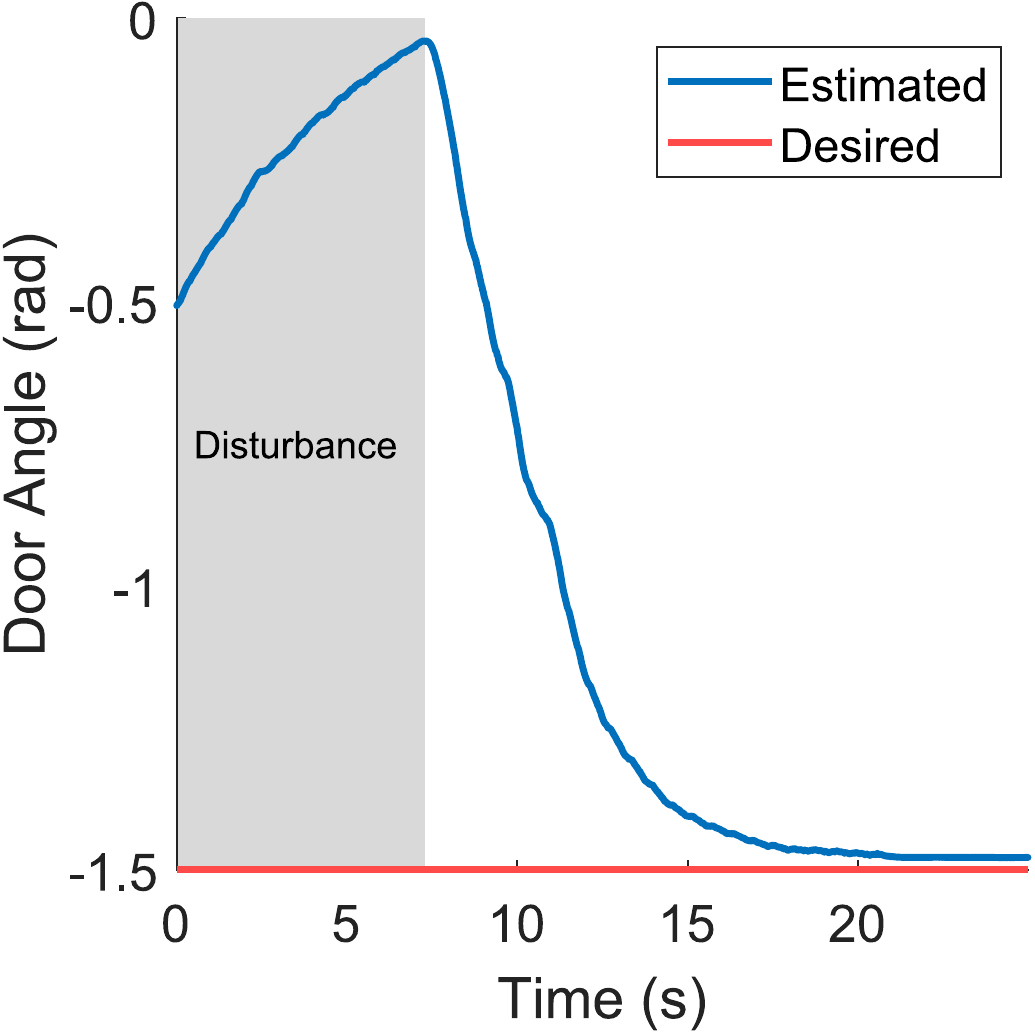}};
    \end{tikzpicture}
        \label{fig:angleforces2}
    \end{subfigure}
\begin{subfigure}[t]{0.5\textwidth}
        \centering
 \begin{tikzpicture}
    \node (A) at (0,0) {\includegraphics[scale=0.26]{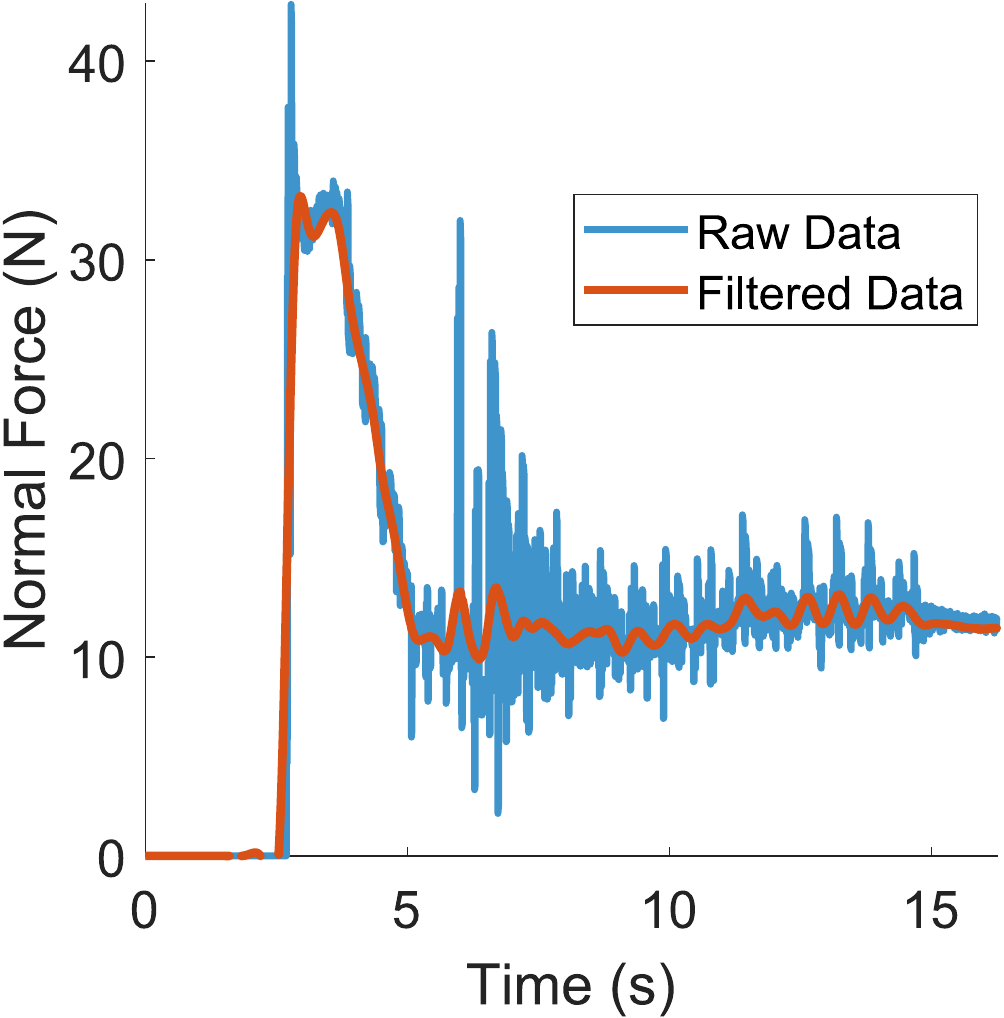}};
    \node[right] (B) at (A.east) {\includegraphics[scale=0.26]{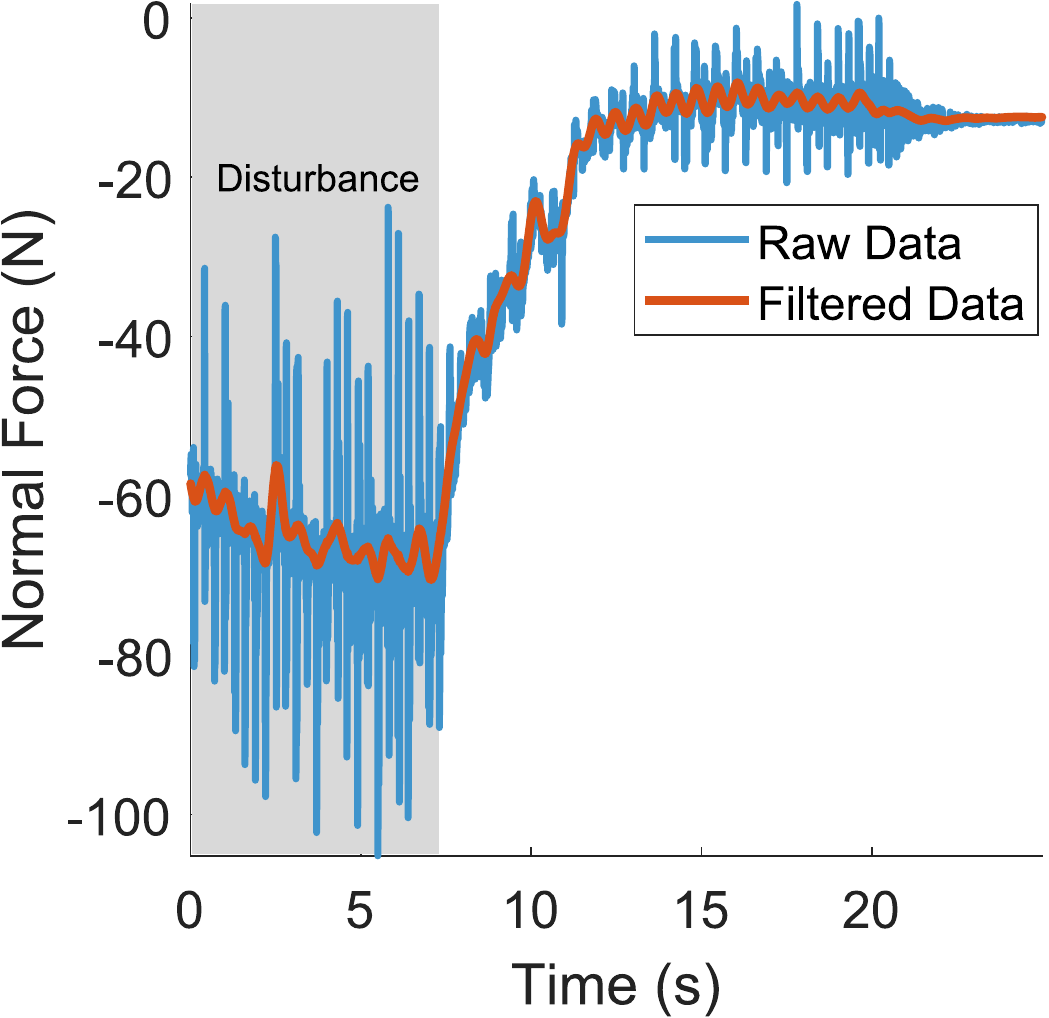}};
    \end{tikzpicture}
        \label{fig:angleforces2}
    \end{subfigure}
    \caption{Plots showing the estimated door angle and the normal force applied to the door for the pushing \textit{(left)} and pulling \textit{(right)} experiments. The door is disturbed during the pulling task.}
    \label{fig:PushDoor}
    \vspace{0mm}
\end{figure}

In this section, we adopt an object-centered perspective in our task-specification, meaning that the objective is defined on the level of the object's state, and the robot's optimal motion/force trajectories are generated accordingly. Therefore, we set $\alpha_1 = 0$, $\alpha_2 = 1$ and $\alpha_3 = 1$ during our next experiments. The first set of examples consists of simulations of the following manipulation tasks (see attached video): Dragging (pulling) a ${10 \ \text{kg}}$ load, dynamically throwing a ${3 \ \text{kg}}$ load, and dynamically pushing a ${10 \ \text{kg}}$ load, all towards a target position. The first one has a continuously closed contact state, while the other two have a closed contact initially, followed by an open-contact after a switching time of $t_s = 0.6 \ \text{s}$ from the start of the manipulation sequence. The load-dragging video highlights the importance of imposing arm joint-torque limits in the MPC formulation. This is essential for avoiding arbitrarily large and unattainable contact forces at any arm configuration. We note that in both cases (with and without torque constraints) the task is achieved; however, the constrained version results in the robot naturally extending its arm closer to a singular configuration to be able to apply the forces required to drag the load. Moreover, we verify the adequacy of the new inequality-handling algorithm through the torque plots corresponding to the object-throwing task (see~\cref{fig:torquePlots}). In fact, the system is operated at the boundary of the feasible region without any algorithmic instability issues. Finally,~\cref{fig:Pushing} shows the frame sequences of the dynamic pushing task, where the loss of contact throughout the manipulation period is displayed. It is worth mentioning that the mode schedule for this task could be easily adapted to include three modes (open - closed - open) where the first transition occurs by having the robot strictly follow a pre-defined trajectory towards the object.
In this final example, we perform real hardware experiments on this paper's central manipulation task, namely opening (pushing and pulling) heavy resistive doors. We model the door -- which is equipped with a door-closer -- as a rotational mass-damper system with a constant recoil torque acting on it. The door angle, which is fed back to the whole-body planner, is estimated by relying on the gripper's position and the door's kinematic parameters. Moreover, due to our planner's inability to inherently account for potential collisions between the robot and the door, we shape a desired robot behavior in the cost function to guide the solver towards collision-free motions. For instance, when pushing the door, we impose a higher penalty on the base's lateral movement to avoid collisions with the door frame. 
The results for the optimal force trajectories and the estimated door angle are presented in~\cref{fig:PushDoor} for the door pushing and pulling tasks. In both scenarios, we observe that the planner discovers physically consistent motion/force plans that drive the door to its desired angle regardless of any model mismatches. The steady-state error could in fact be diminished, and the transient response made faster, by simply increasing the object-state tracking weights and reducing the regularization of the arm's contact forces. To further demonstrate the robustness of our approach, we apply an external disturbance that tends to close the door while the robot is already in the process of pulling it open. During this period, it is clear how the planner still adjusts the contact forces according to the current door angle. When the disturbance vanishes, the robot manages to pull the door to the assigned 90 degree set-point.  
\vspace{0mm}
\subsection{Comparative Study}
The purpose of this section is to further support the importance of encoding the dynamic coupling between the base, arm, and manipulated object in the MPC formulation. To this end, we compare the centroidal dynamics model adopted in this paper to the single rigid body dynamics (SRBD) template model used in previous works \cite{Winkler, Ruben1}, both with and without augmenting the object dynamics. The SRBD model is obtained by assuming a fixed full-body inertia around a nominal robot configuration, and by neglecting the contributions of the joint velocities on the base motion in~\cref{eq:BaseDerivative}. In all four cases the object dynamic effects are incorporated in the tracking controller, and the MPC cost weights are similar. The comparison is based on a dynamic object-lifting task with a displacement of ${1.25 \ \text{m}}$, where various desired lifting times are specified for each case. The times below which the task execution fails are reported in~\cref{table1}, along with the true settling times, and the average computational times of each MPC iteration. We note from the table and from the corresponding videos that including the dynamic effects of the arm and object in the planner allows for a wider range of fast lifting motions. Moreover, we notice a slight decrease in the average computational times when excluding the object's state from the system flow map and/or when using the SRBD approximation. Nevertheless, the best performance was still obtained by the richest model, even if the MPC solver in this case runs at a lower update rate. We note that in this experiment, the failure cases were not caused by any constraint violations, but rather by the SLQ solver's inability to converge to an acceptable solution (i.e., satisfying the assigned tolerances) within the allowable number of function calls. This occurs because at some point, the optimal input trajectory computed at the last MPC iteration leads to a diverging state trajectory during the forward rollout phase of the current iteration. The bigger the discrepancy between the template model and the actual model, the greater the mismatch between the future predicted behavior under the optimal inputs and the measured one, and the more likely it is that the solver fails to converge. Finally, it is worth mentioning that in the case of simply commanding the robot in free-motion with a fixed nominal configuration for the arm and a nominal base orientation, the SRBD model is sufficient. 
\section{CONCLUSION AND FUTURE WORK}
\label{Conclusion}
In this work, we proposed, to the best of our knowledge, the first holistic MPC framework that plans whole-body motion/force trajectories for tasks combining dynamic locomotion and manipulation. The underlying multi-contact optimal control problem is formulated as a constrained and switched system that is solved in real-time with a dedicated SLQ algorithm. We define an extended system model that augments the manipulated-object dynamics to the robot's centroidal dynamics and full kinematics. By that, we are able to account for the coupling effects between the base, limbs, and object in the same planning framework. We demonstrate the effectiveness of our approach on a broad range of test cases with a quadrupedal mobile manipulator performing either free-motion tasks or object-manipulation tasks. We verify that the computed optimal trajectories are physically tractable and respect the system's operational limits as they can be easily tracked by our whole-body controller, and thus easily deployed on the real system. We also validate that the MPC generates solutions fast enough to provide our closed-loop system with robustness properties, which allow it to mitigate the effects of model mismatches or external disturbances. 
One of the natural extensions to this work would be an MPC framework that is aware of the different body geometries and is thereby able to implicitly plan for an optimal contact schedule based on the potential collisions between these bodies. Another interesting direction would be to make the planner adaptive with respect to the object's dynamic properties. This would be done by fusing the current method with online system-identification techniques in order to resolve the parametric uncertainties in the object's model.
\begin{table}[]
\captionsetup{justification=centering}
\centering
\caption{\footnotesize \scshape COMPARATIVE RESULTS FOR 4 TEMPLATE \\ \footnotesize \scshape MODELS DURING THE DYNAMIC LIFTING TASK}
\resizebox{0.95\columnwidth}{!}{%
\begin{tabular}{l|cccc}
                                                                                                  & \begin{tabular}[c]{@{}c@{}}\Large \textbf{Centroidal}\\\Large \textbf{with Object}\end{tabular} & \begin{tabular}[c]{@{}c@{}}\Large\textbf{Centroidal}\\\Large \textbf{without Object}\end{tabular} & \begin{tabular}[c]{@{}c@{}}\Large\textbf{SRBD}\\\Large\textbf{with Object}\end{tabular} & \begin{tabular}[c]{@{}c@{}}\Large\textbf{SRBD}\\\Large\textbf{without Object}\end{tabular}  \\ 
\hline
\multicolumn{1}{c|}{\begin{tabular}[c]{@{}c@{}}\Large Minimum Reference \\ \Large Lifting Time (s)\end{tabular}} & \Large \color{blue} 0.2 \rule{0pt}{2.9ex}                                                              & \Large 1.0   \rule{0pt}{2.9ex}                                                               & \Large 1.5                     \rule{0pt}{2.9ex}                                    & \Large 2.0     \rule{0pt}{2.9ex}                                                        \\
                                                                                                  & \multicolumn{1}{l}{}                                            & \multicolumn{1}{l}{}                                               & \multicolumn{1}{l}{}                                      & \multicolumn{1}{l}{}                                          \\
\multicolumn{1}{c|}{\Large Settling Time (s)}                                                            & \Large \color{blue} 0.77                                                             & \Large 3.64                                                             & \Large 6.75                                                    & \Large 6.90                                                            \\
                                                                                                  & \multicolumn{1}{l}{}                                            & \multicolumn{1}{l}{}                                               & \multicolumn{1}{l}{}                                      & \multicolumn{1}{l}{}                                          \\
\multicolumn{1}{c|}{\begin{tabular}[c]{@{}c@{}}\Large Average MPC \\ \Large Computational Time (ms)\end{tabular}}                                       & \Large 19.9                                                         & \Large 17.2                                                           & \Large 19.2                                                      & \Large \color{blue} 14.5                                                         
\end{tabular}
}
\label{table1}
\vspace{0mm}
\end{table}


\addtolength{\textheight}{0cm}   



\vspace{0mm}
\bibliographystyle{IEEEtran}
\bibliography{bibliography}

\end{document}